\documentclass[preprint,12pt]{elsarticle}

\usepackage{amssymb}
\usepackage{graphicx}
\usepackage{amsmath}
\usepackage{amssymb}
\usepackage{booktabs}

\usepackage[pagebackref,breaklinks,colorlinks]{hyperref}
\usepackage{times}
\usepackage{epsfig}
\usepackage{graphicx}
\usepackage{amsmath}
\usepackage{amssymb}
\usepackage{adjustbox}
\usepackage{tabularx}
 \usepackage{verbatim}
\usepackage{bm}
\usepackage{subcaption}

 \usepackage{algorithm}

\usepackage{algorithmic}%

\usepackage[capitalize]{cleveref}
\crefname{section}{Sec.}{Secs.}
\Crefname{section}{Section}{Sections}
\Crefname{table}{Table}{Tables}
\crefname{table}{Tab.}{Tabs.}

\journal{Nuclear Physics B}

\begin{document}

\begin{frontmatter}



\title{Unsupervised  Domain Adaptation Using Compact Internal Representations}


\author{Mohammad Rostami\\rostamim@usc.edu}

\affiliation{organization={University of Southern California}
}
\begin{abstract}
 A major technique for tackling unsupervised domain adaptation involves mapping data points from both the source and target domains into a shared embedding space. The mapping encoder to the embedding space is trained   such that the embedding space becomes   domain agnostic, allowing a classifier trained on the source domain to generalize well on the target domain. To further enhance the performance of unsupervised domain adaptation (UDA), we develop an additional technique which makes the internal distribution of the source domain more compact, thereby improving the model's ability to generalize in the target domain.
We demonstrate that by increasing the margins between data representations for different classes in the embedding space, we can improve the model performance for   UDA. To make the internal representation more compact, we estimate the internally learned multi-modal distribution of the source domain as Gaussian mixture model (GMM). Utilizing the estimated GMM, we enhance the separation between different classes in the source domain, thereby mitigating the effects of domain shift. We offer theoretical analysis to support outperofrmance of our method. To evaluate the effectiveness of our approach, we conduct experiments on  widely used UDA benchmark UDA  datasets. The results   indicate that our method enhances model generalizability and outperforms existing techniques. \footnote{This paper is based on results partially presented at the 2022 Conference on Lifelong Learning Agents \cite{rostami2022increasing}.}
\end{abstract}







\begin{keyword}
domain adaptation, model generalization, internal distribution
\end{keyword}

\end{frontmatter}



\section{Introduction}

Despite the remarkable advancements in deep learning, deep neural networks still face challenges with generalization when distributional gaps occur during model execution. These gaps result from ``domain shift'' which occurs when there are significant differences between the distribution of data in the training domain (source domain) and the testing domain (target domain). To overcome this issue and ensure effective generalization in the target domain, it is necessary to retrain the neural network using data specifically from the target domain. However, this process can be expensive and time-consuming since it requires manual data annotation and preparing a large training dataset \cite{kim2022did,rostami2018crowdsourcing}. 
Moreover, the process becomes infeasible when the distribution changes continually~\cite{rostami2021cognitively,rostami2021detection,saporta2022multi}, making persistent data annotation necessary.

Unsupervised Domain Adaptation (UDA) is a learning framework aimed at addressing the challenges posed by domain shift when only unlabeled data is available in a target domain. In UDA, the primary strategy involves mapping the labeled data from the source domain and the unlabeled data from the target domain into a shared latent embedding space. This shared space serves as a bridge between the two domains, facilitating knowledge transfer and adaptation ~\cite{rostami2022transfer}.
The objective in UDA is to minimize the distributional discrepancy between the source and target domains in this shared embedding space. By reducing the distance between the distributions, the effects of domain shift can be mitigated, enabling improved generalization in the target domain \cite{long2015learning,he2016deep,stan2022unsupervised,zhang2021adaptive,chen2023activate,li2020sequential}. 
Domain alignment allows for the training of a classifier network in the source domain using data representations obtained from this shared space. Since the data representations in the embedding space capture the common underlying structure between the domains, the classifier network can effectively learn to discriminate between the classes using the shared knowledge. As a result, it demonstrates good generalization capabilities in the target domain, despite being trained exclusively on data from the source domain. By exploiting the information contained in the unlabeled target data through the shared space, UDA provides a powerful mechanism for adapting models to new domains without requiring costly manual annotation efforts.

Most domain adaptation methods have focused on modeling the shared embedding space as the output space of a deep encoder network and training the encoder to enforce domain alignment. Various approaches have been extensively explored to achieve domain alignment. Most methods aim to enforce domain alignment through adversarial learning or direct cross-domain distribution alignment \cite{ganin2015unsupervised,ganin2016domain,tzeng2017adversarial,hoffman2018cycada,bhushan2018deepjdot,pan2019transferrable,rostami2021lifelong,rostami2023concpt,jian2023unsupervised,oza2023unsupervised,westfechtel2024gradual,zhang2024category}.
Adapting generative adversarial networks (GANs) to UDA involves aligning two distributions indirectly within an embedding space. The shared encoder is modeled as a cross-domain feature generator network. By employing a competing discriminator network, which is jointly trained using an adversarial min-max training procedure, the source domain and target domain features are made indistinguishable. 
The generator networks is trained to produce samples that can ``fool'' the discriminator into classifying them as  target domain samples or the source domain samples. When the discriminator is unable the distinguish the correct domain for the samples, the embedding space becomes domain-agnostic.
This alignment procedure effectively aligns the two distributions but adversarial learning requires careful engineering, including setting the initial optimization point, designing the auxiliary networks' architecture, and selecting appropriate hyperparameters to maintain stability. Adversarial learning also has vulnerabilities, such as mode collapse, which can impact its effectiveness. Due to these challenges,  direct probability matching has been explored as an alternative approach.

Direct probability matching focuses on minimizing the distance between two domain-specific distributions directly in the embedding space~\cite{redko2017theoretical}. This approach requires less data engineering if an appropriate probability distribution metric is chosen. However, methods based on probability matching tend to have lower average performance compared to those based on adversarial learning. One reason for this discrepancy is the challenge of measuring distances in higher dimensions, known as the curse of dimensionality. It is difficult to establish meaningful semantic similarities in the input space based solely on small distances between distributions in an embedding space. Thus, the key challenge lies in selecting a suitable probability distribution metric that can effectively measure distances in high dimensions, using empirical samples, and can be optimized efficiently.
Overall, both adversarial learning and direct probability matching offer different approaches to achieve domain alignment in UDA. Adversarial learning requires careful engineering but can be effective in aligning distributions, while direct probability matching requires selecting an appropriate metric that can capture meaningful distances in high-dimensional spaces.

 One promising direction to enhance the general strategy of domain alignment in UDA is to incorporate additional mechanisms that improve the separability of target domain data in the embedding space. Recent advances in UDA primarily stem from the introduction of new secondary mechanisms that enhance performance.
Various approaches have been explored to improve UDA performance by leveraging secondary mechanisms. For instance, Motiian et al. \cite{motiian2017unified} enforce semantic class-consistent alignment of distributions by assuming access to a few labeled target domain data. Chen et al. \cite{chen2019joint} align the covariances of the source and target domains to reduce domain discrepancy.   Li et al. \cite{li2020enhanced} introduce entropy-based regularization into the loss function to leverage the intrinsic structure of the target domain classes.
In this work, we would like to explore the possibility of using internal representations to develop a secondary mechanism to improve UDA.
In supervised classification settings, when training a deep neural network, data representations (i.e., network responses) tend to form distinct class-specific clusters in the final layer of the network. An effective strategy to improve domain alignment is to create larger margins between these class-specific clusters in the embedding space, allowing them to be more separated and distinct. Kim et al. \cite{kim2019unsupervised} propose an approach based on adversarial min-max optimization to increase model generalizability by inducing larger margins. This strategy aligns with recent theoretical findings that demonstrate the benefits of large margin separation in the source domain for improved model generalizability in UDA \cite{dhouib2020margin}.

 In order to enhance the generalizability of models in UDA, we introduce a novel secondary mechanism. Our approach leverages the internal data distribution that emerges in a shared embedding space, which is obtained through pretraining on the source domain. The goal is to increase the margins between different visual class clusters, thereby mitigating the impact of domain shift when adapting to a target domain.
The internally learned distribution in the embedding space exhibits a multimodality behavior. To effectively capture this multimodal nature, we employ a Gaussian Mixture Model (GMM) to parametrize and estimate the internal distribution. By using the GMM, we aim to create larger separations between the class clusters.
To increase the interclass margins, we construct a pseudo-dataset by sampling random instances with confident labels from the estimated GMM. The model is then regularized to push the target domain samples away from the class boundaries using the generated pseudo-dataset. This is achieved by minimizing the distance between the target domain distribution and the distribution of the pseudo-dataset which is more compact than the original source domain internal distribution. The objective is to encourage clear separation between classes in the target domain.
Theoretical analysis supports our method by demonstrating that it minimizes an upper bound for the expected error in the target domain. This theoretical foundation provides insight into the effectiveness of our approach.
To evaluate the performance of our algorithm, we conduct experiments on standard  UDA benchmark datasets. Through these evaluations, we demonstrate that our method is competitive compared to existing approaches. The results showcase the efficacy of our approach in improving model generalizability and addressing the challenges posed by domain shift in UDA.

\section{Related Work}    

 The main challenge in the direct probability matching approach lies in selecting an appropriate probability discrepancy measure that can effectively align the distributions between the source and target domains. The chosen metric should be computationally efficient and possess desirable properties for numerical optimization.
Several probability metrics have been employed for probability matching in UDA. One commonly used metric is the Maximum Mean Discrepancy (MMD), which aligns the means of the two distributions \cite{long2015learning,long2017deep,li2020intelligent,zhang2020discriminative}. Building upon this baseline, Sun et al. \cite{sun2016deep} improve the alignment by considering distribution correlations and leveraging second-order statistics. Other approaches include utilizing the Central Moment Discrepancy \cite{zellinger2016central,chen2020homm} and incorporating Adaptive Batch Normalization \cite{li2018adaptive}.
While aligning lower-order probability moments is straightforward, it tends to overlook discrepancies in higher-order moments. To address this limitation, the Wasserstein distance (WD) has been employed \cite{courty2017optimal,damodaran2018deepjdot,xu2020reliable,el2022hierarchical}. WD incorporates information from higher-order statistics and has been shown by Damodaran et al. \cite{damodaran2018deepjdot} to enhance UDA performance compared to MMD or correlation alignment approaches \cite{long2015learning,sun2016deep}. Hence, WD is a suitable choice for domain alignment.

Unlike more classic probability discrepancy measure such as KL-divergence or JS-divergence, WD exhibits non-vanishing gradients even when the two distributions do not have overlapping supports. Note that at the initial stage of UDA, the two distribution might have minimal overlapping supports due to distributional mismatches. This property makes WD well-suited for solving deep learning optimization problems, which are typically employed in deep learning to optimize objective functions through gradient-based methods.
By leveraging the properties of WD and its compatibility with gradient-based optimization, the direct probability matching approach enables effective distribution alignment in UDA. It provides a principled framework for minimizing the distributional differences between domains, facilitating knowledge transfer and enhancing the generalization capability of models in target domains. 
  Although using WD  leads to improved UDA performance, a downside of using WD is heavy computational load in the general case compared to simpler probability metrics. The high computational load is because WD is defined as a linear programming optimization and does not have a closed-form solution for dimensions more than one.  To account for this constraint, we use the sliced Wasserstein distance (SWD)~\cite{lee2019sliced} which we have previously been used for addressing UDA in different settings~\cite{rostami2019sar,rostami2019deep,gabourie2019learning,stan2021privacy,stan2021unsupervised,stan2021unsupervised,stan2022secure,wu2023unsupervised}. 
  SWD is defined in terms of a closed-form solution of WD in 2D.   It can also be computed fast from empirical   samples. Compared to these works that use SWD for distributional alignment, we develop a secondary mechanism that improves model generalizability.

\begin{figure*}[t!]
    \centering
    \includegraphics[width= \linewidth]{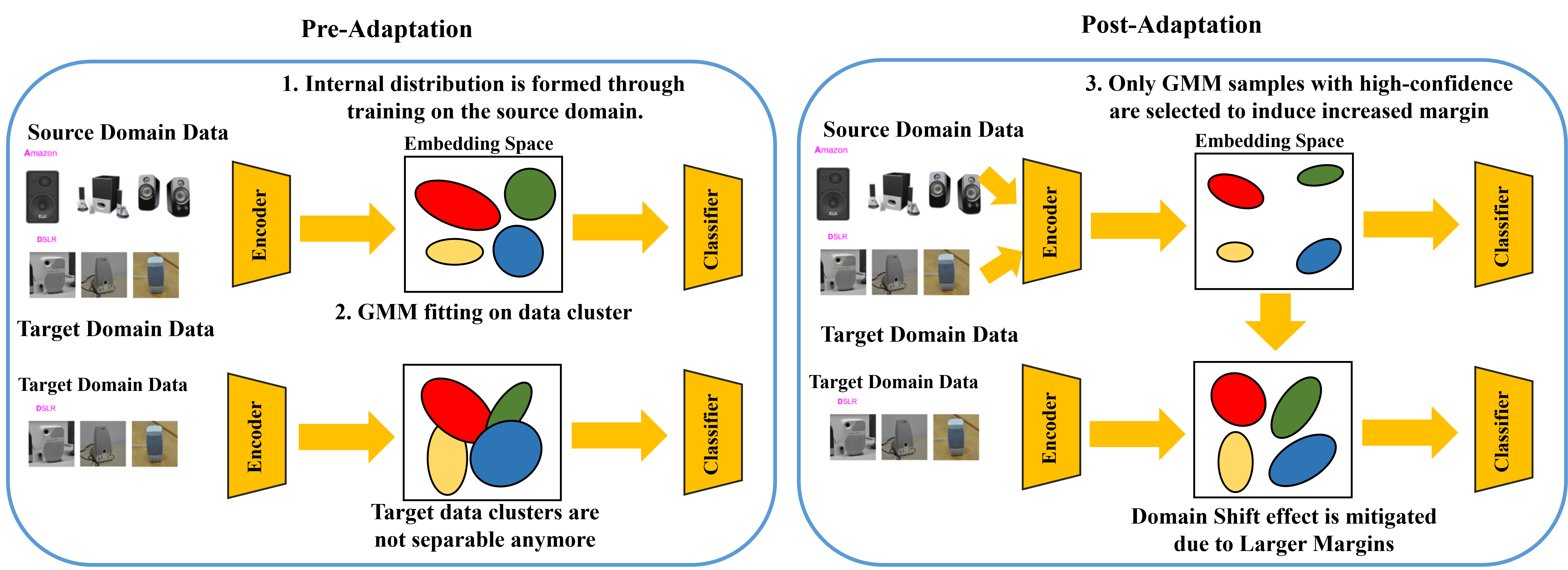}
         \caption{High-level description of the proposed unsupervised domain adaptation algorithm: Initially, a pretrained model on the source domain learns distinct and separable clusters in the embedding space, as shown in the top-left image. However, when this model is applied to the target domain,   the model's performance deteriorates due to reduced class separability in the embedding space, as depicted in the bottom-left image.
To address this issue, our algorithm generates a pseudo-dataset by selecting confident samples from the target domain. This pseudo-dataset is used to induce larger margins between the clusters representing different classes in the embedding space. By making each class cluster more compact, we create additional separation between them. Consequently, the model's generalizability improves, as illustrated in the top-right image. This enhanced generalizability aids in mitigating the effects of domain shift in the target domain, allowing for greater tolerance to domain variations before violating the margins between the classes, as depicted in the bottom-right image.
Our algorithm leverages the pseudo-dataset and the concept of compactness in class clusters to improve the model's performance in the target domain, enabling mitigating the effect of domain shift.}
         \label{figMUDA:1}
\end{figure*}

When training deep neural network classifiers using supervised learning, the goal is to achieve a clear separation between the input data representations in an embedding space, as depicted in Figure~\ref{figMUDA:1}. This embedding space is typically captured by the responses of the network in a higher layer. Through this transformation, the original input distribution is converted into an internal distribution that consists of multiple modes, where each mode corresponds to a specific class. This transformation is crucial for facilitating class discrimination, as the final layer of a classifier network is often a softmax layer that assumes linear separability between classes.
The properties of this internally learned distribution can be effectively utilized to enhance the performance of Unsupervised Domain Adaptation (UDA) methods. An effective approach in UDA focuses on aligning the internal distributions by matching the cluster-specific means for each class across both the source and target domains \cite{pan2019transferrable,chen2019progressive}. This class-aware domain alignment procedure addresses the challenge of class mismatch, where the class distributions in the source and target domains may differ significantly.
By leveraging the learned internal distribution, UDA methods can effectively bridge the gap between different domains and improve the generalization capabilities of the model in the target domain. This alignment process ensures that similar classes from different domains are closer to each other in the embedding space, allowing for better adaptation and transferability of the learned knowledge. By considering the distributional properties of the data, UDA methods can effectively mitigate the challenges posed by domain shift and improve the overall performance and robustness of deep neural network classifiers in real-world applications~\cite{oza2023unsupervised,li2020model,kundu2020universal,stan2023preserving}.

Our goal is to add a regularization component to the shared encoder network in order to maintain the robustness of the internal distribution when the input distribution undergoes perturbations \cite{zhang2020unsupervised}. This is achieved by making the clusters corresponding to different classes more compact in the embedding space. By promoting compactness in the representation of data, we introduce larger margins between classes, which in turn enhances the stability of the model when faced with domain shift. Theoretical studies conducted in few-shot learning scenarios have shown that increasing the margin between classes can effectively reduce the misclassification rate \cite{cao2019theoretical}.
Motivated by these theoretical insights, we propose a novel approach that involves estimating the internal distribution formed in the embedding space using a parametric Gaussian Mixture Model (GMM).
 The GMM allows us to capture the multimodal nature of the distribution and represent it as a weighted combination of Gaussian components. By leveraging this estimated internal distribution, we can induce larger margins between the class-specific distributional modes. This means that the representations of different classes will be more separated, enabling better discrimination and improving the model's ability to generalize to unseen data.
By encouraging margin enhancement through the use of the estimated internal distribution, our approach provides a principled way to address the challenges posed by domain shift. 
  
\section{Problem Statement}  
  
Consider a learning scenario where we have a target domain $\mathcal{T}$ for which we have access to unlabeled data $D_\mathcal{T} = (\bm{X}_\mathcal{T})$, where $\bm{X}_{\mathcal{T}}=[\bm{x}1^t,\ldots,\bm{x}M^t]\in\mathcal{X}\subset\mathbb{R}^{d\times M}$ represents the input data points. Our objective is to train a model that can generalize well on this target domain. However, since we lack annotations for the target domain, using supervised learning to solve this task becomes infeasible. To overcome this limitation, we also assume that we have access to a source domain $\mathcal{S}$ with a labeled dataset $D_\mathcal{S} = (\bm{X}_\mathcal{S},\bm{Y}_\mathcal{S})$, where $\bm{X}_{\mathcal{S}}=[\bm{x}_1^s,\ldots,\bm{x}N^s]\in\mathcal{X}\subset\mathbb{R}^{d\times N}$ and $\bm{Y}_{\mathcal{S}}=[\bm{y}^s_1,...,\bm{y}^s_N]\in \mathcal{Y}\subset\mathbb{R}^{k\times N}$ represent the input data points and their corresponding one-hot labels, respectively. We assume that doth domains share the same set of $k$ semantic classes. This assumption is critical in UDA because it makes the two problems related.

The source and target data points are independently and identically drawn from their respective domain-specific distributions, denoted as $\bm{x}_i^s\sim p_{S}(\bm{x})$ and $\bm{x}_i^t\sim p_{T}(\bm{x})$. However, despite similarities between the two domains, there exists a discrepancy in their distributions, i.e., $p_{T}(\bm{x})\neq p_{S}(\bm{x})$. In the absence of labeled data in the target domain, a naive approach would be to use supervised learning in the source domain for model training. To this end, we consider a family of parameterized functions $f_{\theta}:\mathbb{R}^d\rightarrow \mathcal{Y}$, such as deep neural networks with learnable weight parameters $\theta$. We then search for the optimal model $f_{\theta^*}(\cdot)$ in this family using empirical risk minimization (ERM) on the labeled source domain data:
\begin{equation}
\hat{ \theta}=\arg\min_{\theta}{\hat{e}{\theta}(\bm{X}_{\mathcal{S}},\bm{Y}_{\mathcal{S}},\mathcal{L})}=\arg\min{\theta}{\frac{1}{N}\sum_i \mathcal{L}(f_{\theta}(\bm{x}_i^s),\bm{y}_i^s)},
\end{equation}
 where $\mathcal{L}(\cdot,\cdot)$ represents a suitable point-wise loss function, such as the cross-entropy loss, and the additive terms is the emprical risk.
Ideally, the model trained using ERM would generalize well on the source domain, and due to the transfer of knowledge between the two domains, it would perform better than random guessing on the target domain. However, due to the presence of a domain gap, i.e., $p_{T}(\bm{x})\neq p_{S}(\bm{x})$, the performance of the source-trained model would still be degraded compared to its performance on the source domain.

The above simple solution neglects the potential value of the unlabeled data available in the target domain. In UDA, our objective is to capitalize on the knowledge embedded in the unlabeled target data and the similarities between the two domains in order to enhance the generalization capability of the model trained on the source domain. By benefiting from knowledge transfer, we aim to bridge the gap between the source and target domains by mapping their respective data points into a shared embedding space denoted as $\mathcal{Z}$. This mapping process is facilitated by employing a shared encoder, which is designed to minimize the distribution discrepancy between the domains within the embedding space. The ultimate goal is to minimize the disparity between the domains to the extent that it becomes negligible after the transformation.
 Different UDA algorithms adopt diverse strategies to accomplish this task, leading to a wide range of approaches in the field. The key objective remains finding the shared embedding space that facilitates seamless knowledge transfer between the domains and enables the model to generalize well in the target domain in the presence of domain shift.

To model the shared embedding described above, we can decompose the end-to-end deep neural network $f_\theta(\cdot)$ into two subnetworks: an encoder subnetwork $\phi_{\bm{v}}(\cdot): \mathcal{X}\rightarrow \mathcal{Z}\subset \mathbb{R}^p$ and a classifier subnetwork $h_{\bm{w}}(\cdot): \mathcal{Z}\rightarrow \mathcal{Y}$, where $\bm{v}$ and $\bm{w}$ represent the learnable parameters of the encoder and classifier, respectively. This decomposition can be expressed as $f_\theta = h_{\bm{w}}\circ \phi_{\bm{v}}$, where $\theta=(\bm{w},\bm{v})$ represents the combined parameters.
In line with the condition for good model generalization on the source domain, we assume that the classes are separable in the shared embedding space $\mathcal{Z}$ as a result of supervised pretraining on the source domain (refer to Figure~\ref{figMUDA:1}, left). Note that this assumption is natural because if we cannot train a good model for the source domain, then by extension, we cannot adapt it to work in a target domain with unannotated data. Taking advantage of the target domain data, most UDA frameworks adapt the encoder trained on the source domain in order to match the distributions of both domains in the embedding space $\mathcal{Z}$. In other words, we aim to ensure that $\phi(p_{\mathcal{S}}(\cdot)) \approx\phi(p_{\mathcal{T}}(\cdot))$. By achieving this prperty, the classifier subnetwork can generalize effectively in the target domain, despite being trained solely on the source domain data.
A major class of UDA methods focus on matching the distributions $\phi(p_{\mathcal{S}}(\bm{x}^s))$ and $\phi(p_{\mathcal{T}}(\bm{x}^t))$ by training the encoder $\phi(\cdot)$ to minimize the distance between these two distributions in the embedding space, using a suitable probability distribution metric. The objective is to find an appropriate alignment between the source and target domain distributions in the shared embedding space by minimizing the probability metric in addition to minimizing the ERM loss function on the source domain:
 \begin{equation}
 {
\begin{split}
\hat{\bm{v}},\hat{\bm{w}}=&\arg\min_{\bm{v},\bm{w}} \frac{\lambda}{N}\sum_{i=1}^N \mathcal{L}\big(h_{\bm{w}}(\phi_{\bm{v}}(\bm{x}_i^s)),\bm{y}_i^s\big)+
 D\big(\phi_{\bm{v}}(p_{\mathcal{S}}(\bm{X}_{\mathcal{S}})),\phi_{\bm{v}}(p_{\mathcal{T}}(\bm{X}_{\mathcal{T}}))\big),
\end{split}
}
\label{eq:smallmainPrMatch}
\end{equation}  
where, $D(\cdot,\cdot)$ represents a probability discrepancy metric, and $\lambda$ serves as a trade-off parameter balancing the empirical risk and the domain alignment terms. Multiple probability discrepancy metrics exist, and various UDA methods have been developed based on different choices for $D(\cdot,\cdot)$. For this particular study, the selected metric for $D(\cdot,\cdot)$ is SWD.
For details on definition and computation of SWD, please refer to the Appendix.
In short, we use SWD because it can be computed efficiently from empirical samples of two distributions.

\section{Proposed Algorithm for Increased Interclass Margins}

To increase the margins between classes, we leverage the internally learned distribution. Based on the aforementioned rationale, this distribution takes the form of a multi-modal distribution denoted as $p_J(\cdot)=\phi_{\bm{v}}(p_{\mathcal{S}}(\cdot))$, where $\phi_{\bm{v}}$ represents a mapping function applied to the source domain distribution $p_{\mathcal{S}}(\cdot)$ (refer to Figure~\ref{figMUDA:1}, left). It is important to note that the formation of a multi-modal distribution is not guaranteed, but when we model the embedding space using network responses just prior to the softmax layer, it becomes a prerequisite for our model to learn the source domain. This assumption is not a limitation for our method because it is a pre-condition in UDA.

The means of the distributional modes (clusters) correspond to the concept of ``class prototypes'' which has been explored in previous works for achieving class-consistent domain alignment \cite{pan2019transferrable,chen2019progressive}. The margins between classes can be observed in Figure~\ref{figMUDA:1} (left-top) as the geometric distances between the boundaries of the modes in the internally learned distribution $\phi_{\bm{v}}(p_{\mathcal{S}}(\cdot))$. Conversely, domain shift occurs when the internal distribution for the target domain $\phi_{\bm{v}}(p_{\mathcal{T}}(\cdot))$ deviates from the source-learned internal distribution $\phi_{\bm{v}}(p_{\mathcal{S}}(\cdot))$, resulting in increased overlap between class clusters in the target domain (Figure~\ref{figMUDA:1}, left-bottom). In other words, the effect of domain shift in the input space manifests as a crossing of boundaries between certain class clusters.

Our goal is to develop a mechanism that encourages target domain data samples to move away from the interclass margins and towards the class means. This process is visualized in Figure~\ref{figMUDA:1} (right-top). By achieving this property, we aim to enhance the robustness of our model against domain shift in the input space, as shown in Figure~\ref{figMUDA:1} (right-bottom). The key idea is to make the class clusters in the embedding space more compact for the source domain, which in turn allows for greater variability in the target domain.
To implement this idea, we begin by estimating the internally learned distribution in the latent space $\mathcal{Z}$. We use a parametric Gaussian Mixture Model (GMM) distribution to approximate this distribution. The GMM distribution captures the multi-modal nature of the internal distribution, representing different modes that correspond to class prototypes and interclass margins.
The GMM distribution is defined as follows:
\begin{equation}
{
p_J(\bm{z})=\sum_{j=1}^k \alpha_j
\mathcal{N}(\bm{z}|\bm{\mu}_j,\bm{\Sigma}_j).
}
\end{equation}  
In our estimation of the internally learned distribution using a Gaussian Mixture Model (GMM), we assign the mean vectors $\bm{\mu}_j$ and covariance matrices $\bm{\Sigma}_j$ to each mode, while $\alpha_j$ represents the mixture weights. This probability distribution is an appropriate model for capturing the internal probability distribution because we have prior knowledge of the number of modes, denoted as $k$.
As a result, estimating the GMM parameters becomes simpler compared to the general GMM estimation problem, which typically requires iterative and time-consuming procedures such as expectation maximization (EM) \cite{moon1996expectation}. Since we have labeled source data points and know the value of $k$, we can specifically estimate $\bm{\mu}_j$ and $\bm{\Sigma}_j$ as the parameters of $k$ independent Gaussian distributions. The mixture weights $\alpha_j$ can be easily computed using a Maximum a Posteriori (MAP) estimate.
Let $\bm{S}_j$ denote the set of training data points belonging to class $j$ in the source domain, i.e., $\bm{S}_j={(\bm{x}_i^s,\bm{y}i^s)\in \mathcal{D}{\mathcal{S}}|\arg\max\bm{y}_i^s=j }$. The MAP estimation for the GMM parameters can be computed as follows:
\begin{equation}
{
\begin{split}
&\hat{\alpha}_j = \frac{|\bm{S}_j|}{N},\hspace{2mm}\hat{\bm{\mu}}_j = \sum_{(\bm{x}_i^s,\bm{y}_i^s)\in \bm{S}_j}\frac{1}{|\bm{S}_j|}\phi_v(\bm{x}_i^s),\hspace{2mm} \\& \hat{\bm{\Sigma}}_j =\sum_{(\bm{x}_i^s,\bm{y}_i^s)\in \bm{S}_j}\frac{1}{|\bm{S}_j|}\big(\phi_v(\bm{x}_i^s)-\hat{\bm{\mu}}_j\big)^\top\big(\phi_v(\bm{x}_i^s)-\hat{\bm{\mu}}_j\big).
\end{split}
}
\label{eq:MAPestICLR}
\end{equation}

Compared to the computationally complex procedure for EM  \cite{roweis1998algorithms}, the MAP estimation   in Eq.\eqref{eq:MAPestICLR} does not introduce a significant computational burden. The computational complexity of Eq.\eqref{eq:MAPestICLR} is manageable, especially when considering a balanced source domain dataset.
In the case of a balanced source domain dataset, determining whether data points $\bm{x}_i^s$ belong to $\bm{S}_j$ to compute $\alpha_j$ can be achieved with a computational complexity of $O(N)$. Here, $N$ represents the number of data points. Computing the mean vectors $\bm{\mu}_j$ has a computational complexity of $O(NF/k)$, where $F$ denotes the dimension of the embedding space. The computation of covariance matrices $\bm{\Sigma}_j$ has a complexity of $O(F(\frac{N}{k})^2)$. Given that there are $k$ classes, the total computational complexity for estimating the GMM distribution is approximately $O(\frac{FN^2}{k})$.
If we assume that $O(F)\approx O(k)$, which is a reasonable assumption when modeling the embedding space using network responses in the final layer, the total computational complexity becomes $O(N^2)$. This computational overload is relatively small when compared to the computational complexity of one epoch of back-propagation, which needs to be performed multiple times during training. Since deep neural networks typically have a significantly larger number of weights compared to the number of data points $N$, we conclude that GMM estimation is not a demanding procedure.

We utilize the estimated GMM distribution to introduce larger interclass margins and improve the solution obtained through Eq.\eqref{eq:smallmainPrMatch}. Our aim is to create repulsive biases from the class margins, as depicted in Figure\ref{figMUDA:1} (top-right). To accomplish this, we first generate a pseudo-dataset in the embedding space with confident labels. This pseudo-dataset is denoted as $\mathcal{D}{\mathcal{P}}=(\textbf{Z}{\mathcal{P}},\textbf{Y}{\mathcal{P}})$ and is created by randomly sampling from the estimated GMM distribution. Here, $\bm{Z}{\mathcal{P}}=[\bm{z}1^p,\ldots,\bm{z}{N_p}^p]\in\mathbb{R}^{p\times N_p}$ represents the set of generated embedding vectors, $\bm{Y}{\mathcal{P}}=[\bm{y}^p_1,...,\bm{y}^p{N_p}]\in \mathbb{R}^{k\times N_p}$ which contains the corresponding pseudo-labels, and $\bm{z}_i$ is randomly drawn from $\bm{z}_i^p\sim \hat{p}_J(\bm{z})$.
To ensure that these generated samples are located away from the margins and close to the respective class means in the embedding space, we pass all the initially drawn samples through the classifier subnetwork. We only include the samples for which the classifier exhibits a confidence level higher than a predefined threshold, denoted as $0<\tau<1$. This procedure helps us avoid including samples that are in proximity to the class margins. The selected samples will be part of the final pseudo-dataset $\mathcal{D}_{\mathcal{P}}$.
By applying these steps, we ensure that the selected samples in $\mathcal{D}_{\mathcal{P}}$ are positioned away from the margins and closer to their corresponding class means in the embedding space, thus inducing repulsion from the interclass margins.
\begin{equation}
{
\begin{split}
\mathcal{D}_{\mathcal{P}}=\bigg\{&(\bm{z}_i^p,\bm{y}^p_i)|   \bm{z}_i^p\sim \hat{p}_J(\bm{z}),  \max\{ h(\bm{z}_i^p)\}>\tau, \bm{y}^p_i=\arg\max_i\{ h(\bm{z}_i^p)\}\bigg\}.
\end{split}
}
\label{eq:pesudosamples}
\end{equation}    
The process of selecting samples based on threshold values close to one, i.e., $\tau\approx 1$, indicates that the samples in the pseudo-dataset are located in the vicinity of the class means (as illustrated in Figure~\ref{figMUDA:1} (right-top)). Consequently, the margins between the empirical data clusters in the generated pseudo-dataset are larger compared to the empirical clusters of the source domain data in the embedding space. In other words, these samples represent a more compact representation in the embeddign space. We leverage this characteristic of the pseudo-dataset to update Eq.~\eqref{eq:smallmainPrMatch} and induce larger margins:
\begin{equation}
{
\begin{split}
\hat{\bm{v}},\hat{\bm{w}}=&\arg\min_{\bm{v},\bm{w}}\Big\{ \frac{\lambda}{N}\sum_{i=1}^N \mathcal{L}\big(h_{\bm{w}}(\phi_{\bm{v}}(\bm{x}_i^s)),\bm{y}_i^s\big) +\frac{\lambda}{N_p}\sum_{i=1}^{N_p} \mathcal{L}\big(h_{\bm{w}}(\bm{z}_i^p),\bm{y}_i^p\big) +\\&  \hat{D}\big(\phi_{\bm{v}}(\bm{X}_{\mathcal{T}}),\bm{X}_{\mathcal{P}} \big) +  \hat{D}\big(\phi_{\bm{v}}(\bm{X}_{\mathcal{S}}), \bm{X}_{\mathcal{P}} \big)\Big\} ,
\end{split}
}
\label{eq:mainPrMatchICLR}
\end{equation}  
where, $\hat{D}(\cdot,\cdot)$ represents the empirical SWD   metric. In Equation~\eqref{eq:mainPrMatchICLR}, we have four terms that contribute to the optimization objective.
The first and second terms correspond to the ERM objective functions for the source dataset and the pseudo-dataset, respectively. These terms aim to maintain discriminative features in the embedding space for both the source and pseudo-domains, ensuring that the learned representations capture the characteristics of their respective domains.
The third term serves as an alignment term, matching the source domain distribution with the empirical distribution of the pseudo-dataset. This alignment helps to align the source domain with the pseudo-domain, enabling the model to generalize well across both domains.
The fourth term also acts as an alignment term, but it focuses on matching the target domain distribution with the empirical distribution of the pseudo-dataset. Since the pseudo-dataset possesses larger interclass margins, aligning the target domain with this distribution helps increase the margins in the target domain as well, enhancing the model's generalizability in the presence of domain shift.
These terms collectively contribute to increasing the interclass margins, as we discussed earlier, leading to improved model generalizability and adaptation to the target domain.

Our proposed algorithm, named Increased Margins for Unsupervised Domain Adaptation (IMUDA), is visualized and presented in Figure~\ref{figMUDA:1} and Algorithm~\ref{algSAforUDA}, respectively. Figure~\ref{figMUDA:1} provides a visual representation of the concepts and steps involved in IMUDA, illustrating the impact of increased interclass margins on the feature representations visually. The algorithm outlines the steps involved in leveraging the estimated GMM distribution, constructing the pseudo-dataset, and incorporating the alignment terms into the optimization objective to enhance the adaptation process.

 \begin{algorithm}[t]
\caption{IMUDA\label{IJCAI2021Alg}} 
 { 
\begin{algorithmic}[1]
 
\STATE \textbf{Input:}   The datasets $\mathcal{D}_{\mathcal{S}}=(\bm{X}_{\mathcal{S}},  \bm{Y}_{\mathcal{T}})$, $\mathcal{D}_{\mathcal{T}}=(\bm{X}_{\mathcal{S}} )$
\STATE \hspace{4mm}\textbf{Pretraining on the Source Domain:}
\STATE \hspace{4mm} $\hat{ \theta}_0=(\hat{\bm{w}}_0,\hat{\bm{v}}_0)  =\arg\min_{\theta}\sum_i \mathcal{L}(f_{\theta}(\bm{x}_i^s),\bm{y}_i^s)$
\STATE \hspace{2mm}  \textbf{GMM Estimation:}
\STATE \hspace{4mm} Use \eqref{eq:MAPestICLR} and estimate GMM paramters $\alpha_j, \bm{\mu}_j,$   $\Sigma_j$
\STATE \textbf{Domain Adaptation}: 
\STATE \hspace{2mm} \textbf{Pseudo-Dataset Generation:} 
\STATE \hspace{4mm} Generate $\mathcal{\hat{D}}_{\mathcal{P}}$ based on ~\eqref{eq:pesudosamples}
\FOR{$itr = 1,\ldots, ITR$ }
\STATE draw random data batches from $\mathcal{ D}_{\mathcal{S}}$, $\mathcal{ D}_{\mathcal{T}}$, and $\mathcal{ D}_{\mathcal{P}}$ 
  and update the model based on   ~\eqref{eq:mainPrMatchICLR}
\ENDFOR
\end{algorithmic}}
\label{algSAforUDA}
\end{algorithm} 

\section{Theoretical Analysis}

We analyse IMUDA in a standard PAC-learning setting~\cite{shalev2014understanding}. In short, our goal is to prove that IMUDA algorithm minimizes an upperbound for the expected error on the target domain. We define the hypothesis space $\mathcal{H}$ as $\mathcal{H} = \{h_{\bm{w}}(\cdot)|h_{\bm{w}}(\cdot):\mathcal{Z}\rightarrow \mathbb{R}^k, \bm{v}\in \mathbb{R}^V\}$ be
 the hypothesis space. This space includes all the classifiers that can be represented by the classifer network. Also, let $\hat{\mu}_{\mathcal{S}}=\frac{1}{N}\sum_{n=1}^N\delta(\phi_{\bm{v}}(\bm{x}_n^s))$ and $\hat{\mu}_{\mathcal{T}}=\frac{1}{M}\sum_{m=1}^M\delta(\phi_{\bm{v}}(\bm{x}_m^t))$ be the empirical source and the empirical target distributions in $\mathcal{Z}$. Similarly, let $\hat{\mu}_{\mathcal{P}}=\frac{1}{N_p}\sum_{q=1}^{N_p}\delta(\bm{z}_n^q)$ be the empirical internal distribution. These distributions estimate the true distribution which are unkown.
  We denote the expected error for $h(\cdot)_{\bm{w}}\in\mathbb{H}$ on the source and the target domains by $e_{\mathcal{S}}({\bm{w}})$ and  $e_{\mathcal{T}}({\bm{w}})$.  Also, let $h_{\bm{w}^*}$ be the optimal joint-trained model, i.e., $e_{\mathcal{C}}(\bm{w}^*)$, i.e. $\bm{w}^*= \arg\min_{\bm{w}} e_{\mathcal{C}}(\bm{w})=\arg\min_{\bm{w}}\{ e_{\mathcal{S}}({\bm{w}})+  e_{\mathcal{T}}({\bm{w}})\}$.  Finally, note that we have $\tau =  \mathbb{E}_{\bm{z}\sim \hat{p}_{J}(\bm{z})}(\mathcal{L}(h(\bm{z}),h_{\hat{\bm{w}}_0}(\bm{z})) $ because only samples with confident predicted labels are included in the generated pseudo-dataset. We offer the following theorem to justify why our algorithm is effective.
  

\textbf{Theorem 1}: Consider a UDA problem involving a source and a target domain and use algorithm~\ref{IJCAI2021Alg} for model adaptation. Then, the following inequality  holds:
\begin{equation}
\small
\begin{split}
e_{\mathcal{T}}\le & e_{\mathcal{S}} +\alpha SW(\hat{\mu}_{\mathcal{S}},\hat{\mu}_{\mathcal{P}})^\beta+\alpha SW(\hat{\mu}_{\mathcal{T}},\hat{\mu}_{\mathcal{P}})^\beta+(1-\tau)+e_{\mathcal{C'}}(\bm{w}^*)\\&+\sqrt{\big(2\log(\frac{1}{\xi})/\zeta\big)}\big(\sqrt{\frac{1}{N}}+\sqrt{\frac{1}{M }} \big),
\end{split}
\label{eq:theroemforPLnips}
\end{equation}    
where $W(\cdot,\cdot)$ denotes the WD distance and   $\xi$ is a constant which depends on the loss function  $\mathcal{L}(\cdot)$ characteristics,   $\alpha$ is a constant and  $\beta=(2(d+1))^{-1}$.  

\textbf{Proof:}    We base our proof on the applicability of using optimal transport for UDA~\cite{redko2017theoretical}.   We first state the following theorem   developed for UDA when we use the optimal transport for domain matching.

\textbf{Theorem 2~\cite{redko2017theoretical}}: Under the assumptions described   for UDA, then for any $d'>d$ and $\zeta<\sqrt{2}$, there exists a constant number $N_0$ depending on $d'$ such that for any  $\xi>0$ and $\min(N,M)\ge  N_0\max(\xi^{-(d'+2)},1)$ with probability at least $1-\xi$ for all $f_\theta$, the following holds:
\begin{equation}
\begin{split}
e_{\mathcal{T}}\le & e_{\mathcal{S}} +W(\hat{\mu}_{\mathcal{T}},\hat{\mu}_{\mathcal{S}})+e_{\mathcal{C}}(\theta^*)+  \sqrt{\big(2\log(\frac{1}{\xi})/\zeta\big)}\big(\sqrt{\frac{1}{N}}+\sqrt{\frac{1}{M}}\big),
\end{split}
\label{eq:theroemfromcourty}
\end{equation} 
 where $\zeta\in \mathbb{R}$ is constant.

  We use Theorem 2 but since we benefit from SWD rather than optimal transport and rely on pseudo-labeling, it is not directly applicable to our algorithm.   To extend Theorem 2 to our algorithm, we rely on the following relationship between the optimal transport and SWD~\cite{bonnotte2013unidimensional}:
  \begin{eqnarray}
SW(p_X,p_Y)\leq W(p_X,p_Y) \leq \alpha SW(p_X,p_Y)^\beta.
\label{eq:inequalities}
\end{eqnarray}
 
 Building the above two results, we prove Theorem 1.

Considering a minimum confidence threshold of $\tau$ for the pseudo-labeled data points, we can conclude that the probability of the pseudo-label being incorrect is at most $1-\tau$. To quantify the disparity between the error based on the true labels and the pseudo-labels assigned to a specific data point, we can express it as follows:
 \begin{equation}
\begin{split}
  |\mathcal{L}_{ce}(f_\theta(\bm{x}^t_i),\bm{y}^t_i)- \mathcal{L}_{ce}(f_\theta(\bm{x}^t_i),\hat{\bm{y}}_i^{t})|= \begin{cases}
    0, & \text{if $\bm{y}^t_i=\hat{\bm{y}}_i^{t}$}.\\
    1, & \text{otherwise}.
  \end{cases}
\end{split}
\label{eq:theroemforPLproof}
\end{equation}    
We use Jensen's inequality and apply the expectation operator on both sides of Eq.~\eqref{eq:theroemforPLproof}:
\begin{equation}
\begin{split}
& 
|e_{\mathcal{PL}}-e_{\mathcal{T}}|\le\mathbb{E}\big(|\mathcal{L}(f_\theta(\bm{x}^t_i),\bm{y}^t_i)- \mathcal{L}(f_\theta(\bm{x}^t_i),\hat{\bm{y}}_i^{t})|\big)\le (1-\tau).
\end{split}
\label{eq:theroemforPLproofexpectation}
\end{equation}    
Now we use Eq.~\eqref{eq:theroemforPLproofexpectation} and  deduce the following:
\begin{equation}
\begin{split}
&e_{\mathcal{S}}+e_{\mathcal{T}}=e_{\mathcal{S}}+e_{\mathcal{T}}+e_{\mathcal{PL}}-e_{\mathcal{PL}}\le  
e_{\mathcal{S}}+e_{\mathcal{PL}}+|e_{\mathcal{T}}-e_{\mathcal{PL}}|\le \\&
e_{\mathcal{S}}+e_{\mathcal{PL}}+(1-\tau).
\end{split}
\label{eq:theroemforPLprooftrangleinq}
\end{equation}    
 Note that  Eq.~\eqref{eq:theroemforPLprooftrangleinq} holds for any model with a given parameter $\theta$. So, it holds for the joint optimal parameter $\theta^*$  which concludes:
\begin{equation}
\begin{split}
e_C(\theta^*)\le e_{C'}(\theta)+(1-\tau),
\end{split}
\label{eq:theroemforPLprooftartplerror}
\end{equation}    
where $e_{C'}(\theta)$ is the expected error of the model trained on the combination of the target and pseudo-labeled dataset. Now in Theorem 2, let the pseudo-labeled data points be used as the target domain dataset, applying the triangular inequality on WD term first, then applying Eq.~\eqref{eq:theroemforPLprooftartplerror} and Eq.~\eqref{eq:inequalities}  on Eq.\eqref{eq:theroemfromcourty},  Theorem 1 follows.

We utilize Theorem~1 to justify how the IMUDA algorithm can enhance model generalization on the target domain. By comparing Equation\eqref{eq:theroemforPLnips} and Equation~\eqref{eq:mainPrMatchICLR}, we observe that the IMUDA algorithm minimizes an upper bound of the expected error on the target domain.
The first three terms in Equation~\eqref{eq:theroemforPLnips} are directly included in the objective function of Equation~\eqref{eq:mainPrMatchICLR}. The first term, which corresponds to the target expected error $e_{\mathcal{T}}$, is minimized through the alignment of the source and target domain distributions in the embedding space. The second term represents the source expected error $e_{\mathcal{S}}$, which is reduced due to the supervised training on the source domain samples. The third term, which captures the discrepancy between the source and target domains, is minimized through the alignment terms in the optimization problem of Equation~\eqref{eq:mainPrMatchICLR}.

The fourth to sixth terms in Equation~\eqref{eq:theroemforPLnips} are constant terms that provide conditions under which our algorithm can work effectively. The term $(1-\tau)$ becomes small when we set the threshold $\tau$ close to 1, implying that the pseudo-dataset is generated with confident labels, as we do in IMUDA. The term $e_{C'}(\bm{w}^*)$ will be small if the two domains are related, such as sharing the same classes, and a model trained jointly on both domains can achieve good performance. This term indicates that the alignment of distributions in the embedding space must be a feasible task for our algorithm to be effective. The last term in Equation~\eqref{eq:theroemforPLnips} is a constant term that, similar to most learning algorithms, becomes negligible when we have large source and target datasets.
In conclusion, if the domains are related and the conditions mentioned above hold, the IMUDA algorithm minimizes an upper bound of the expected error on the target domain.  
  
 \section{Empirical Validation}
We implement our algorithm and demonstrate that it is effective using standard UDA benchmark datasets. Our implementation is available. We offer comparison results to demonstrate that our method is competitive as well as analytic and ablative experiments to provide a deeper insight about our method.

 \subsection{Datasets and Tasks}
 We validate our method on four standard UDA benchmarks.

 \textbf{Digit recognition   tasks:} The domains used in this benchmark are MNIST ($\mathcal{M}$), USPS ($\mathcal{U}$), and SVHN ($\mathcal{S}$) digit recognition datasets. Consistent with previous research, we focus on three specific digit recognition UDA tasks: $\mathcal{M}\rightarrow \mathcal{U}$, $\mathcal{U}\rightarrow \mathcal{M}$, and $\mathcal{S}\rightarrow \mathcal{M}$. To ensure consistency in the input sizes, we resized the images from the SVHN dataset to dimensions of $28\times 28$, matching the size of the MNIST and USPS datasets.

\textbf{Office-31 Detest:}  this benchmark is a widely recognized UDA dataset, known for its 31 distinct visual classes and a total of 4,652 images. The benchmark comprises three domains: Amazon ($\mathcal{A}$), Webcam ($\mathcal{W}$), and DSLR ($\mathcal{D}$). The benchmark allows for the definition of six UDA tasks, created by pairing the three domains in various combinations.

\textbf{ImageCLEF-DA Dataset:} The benchmark  consists of 12 classes that are shared among three visual recognition datasets: Caltech-256 ($\mathcal{C}$), ILSVRC 2012 ($\mathcal{I}$), and Pascal VOC 2012 ($\mathcal{P}$). Each class in the dataset contains an equal number of images, with 50 images per class or 600 images per domain. This benchmark is unique in that it provides a balanced distribution of images across domains and classes, unlike the Office-31 dataset where domain and class sizes can vary. Similar to the Office-31 dataset, the shared classes in this benchmark allow for the definition of six distinct UDA tasks.


\textbf{VisDA-2017:}  The objective of this benchmark is to develop a model that can effectively learn from synthetic images and generalize well to real-world images. The benchmark involves training the model on samples from a synthetic domain and then adapting it to perform well on the real image domain. The synthetic images are generated using 3D models of various objects, and different lighting conditions are applied to create variations across the 12 classes. The dataset used for training consists of a larger collection of 280,000 images, providing a diverse set of examples for the model to learn from.

We see that these benchmarks are diverse and quite different from each other. Competitive performance on all them can serve as a justification that the proposed UDA algorithm would be a parasitically beneficial algorithm.

\subsection{Backbone Structure and Evaluation Protocol:}

To ensure fair comparisons with existing works, we adopt the network structures commonly used in the literature for each specific dataset. For the digit recognition tasks, we employ the VGG16 network as the backbone model. 
In the case of the Office-31 and ImageCLEF-DA datasets, we utilize the ResNet-50 network architecture as the backbone. This network configuration is widely used in UDA research for these particular datasets, enabling us to make meaningful comparisons with existing methods.
For the VisDa2017 dataset, we opt for the ReNet-101 network as the backbone. 
All backbone models are pretrained on the ImageNet dataset, which provides a strong foundation for feature extraction. The output of the backbone models is then passed through a hidden layer with a size of 128. Following the hidden layer, the number of units in the last layer is set according to the number of classes in each specific dataset. The last layer is implemented as a softmax layer, producing class probabilities. The features extracted before the softmax layer constitute the embedding space $\mathcal{Z}$, which serves as the basis for our UDA approach.
 
We use the cross-entropy loss as the discrimination loss.  At each   training epoch, we computed the combined loss function on the training split of datasets of both domains and stopped training when the training loss function became stable.  We used Keras for implementation and   ADAM optimizer to solve the UDA optimization problems. We have used 100 projection to compute the SWD loss. We tune the learning rate for each dataset such that the training loss function reduces smoothly.  We have run our code on a cluster node equipped with 4 Nvidia Tesla P100-SXM2 GPU's.   We used the  classification rate on  the  testing set to measure performance of the algorithms.   

 To establish a baseline, we first evaluate the performance of the source-trained model on the target domain, referred to as ``Source Only''. This baseline measurement allows us to compare the performance improvements achieved through model adaptation. We then apply the IMUDA algorithm to adapt the model and assess its performance on the target domain.
In our experimental results, we provide the average classification rate along with the standard deviation on the target domain. To ensure statistical robustness, we conduct ten randomly initialized runs for all datasets, except for VisDA2017, where we report the best result achieved.
Based on our theorem, we set the confidence threshold $\tau$ to 0.95. This value ensures that only pseudo-labeled data points with high confidence are included in the adaptation process. Additionally, we set the regularization parameter $\lambda$ to $10^{-2}$. The rationale behind selecting these specific values will be further explored and discussed.

  To ensure a comprehensive comparison against existing UDA methods, we carefully selected a subset of methods that represent both pioneer and recent works in the field. This selection allows us to highlight the advancements made over the pioneer methods and present a snapshot of the recent progress in UDA.In our selection process, we considered methods that have reported results on the majority of benchmarks used in our experiments. This criterion ensures that the selected methods are applicable to a wide range of scenarios and datasets. We included UDA methods based on adversarial learning, such as GtA \cite{sankaranarayanan2018generate}, DANN \cite{ganin2016domain}, 
ADDA \cite{tzeng2017adversarial}, MADA \cite{pei2018multi}, CDAN \cite{long2018conditional}, DMRL \cite{wu2020dual}, DWL \cite{xiao2021dynamic}, HCL \cite{huang2021model},   CGDM \cite{du2021cross}, and ALSDA~\cite{mei2023automatic}. These methods leverage adversarial training techniques to align the source and target domains.
Additionally, we included methods that focus on direct distribution matching, such as DAN \cite{long2015learning}, DRCN \cite{ghifary2016deep}, RevGrad \cite{ganin2014unsupervised}, JAN \cite{long2017deep}, JDDA \cite{chen2019joint}, CADA-P \cite{kurmi2019attending}, ETD \cite{li2020enhanced}, MetaAlign \cite{wei2021metaalign},   FixBi \cite{na2021fixbi},   CAF~\cite{xie2022collaborative},   TDUDA~\cite{huang2022balancing},   PICSCS~\cite{li2023pseudo}, and COT~\cite{liu2023cot}. These methods aim to reduce the distribution discrepancy between the source and target domains using various matching techniques.

For each benchmark dataset where we report our results, we included the performance of the aforementioned methods if their original papers reported results on that particular dataset. In our result tables, we highlight the best performance achieved among all methods using bold font. The tables are structured to present the pre-adaptation baseline performance in the first row, followed by UDA methods based on adversarial learning, and finally UDA methods based on direct distribution matching. Our own results are presented in the last rows of the tables.
By selecting a diverse set of representative UDA methods, we aim to provide a comprehensive comparison and demonstrate the effectiveness of our IMUDA algorithm in improving performance on various benchmarks compared to existing state-of-the-art approaches.

\subsection{Results}

In Table~\ref{table:tabDA1}, we present the performances of various methods, including our IMUDA algorithm, on the three digit recognition tasks. Upon analyzing the results, we observe that IMUDA achieves competitive performance in these tasks. Interestingly, we also note that methods such as ETD, JDDA, and DWL, which incorporate secondary alignment mechanisms, demonstrate competitive performances as well. This observation indicates that the inclusion of secondary alignment mechanisms is crucial for enhancing the performance of current UDA methods, considering the performance levels achieved by existing UDA approaches.

The competitiveness of IMUDA in these tasks showcases its effectiveness in improving model generalization and adaptation to target domains. Furthermore, the strong performance of methods utilizing secondary alignment mechanisms highlights the importance of incorporating additional alignment strategies beyond traditional domain alignment techniques.
These findings emphasize the need for further exploration and development of UDA methods with enhanced alignment mechanisms. By leveraging secondary alignment mechanisms, we can potentially enhance the capability of UDA models to bridge the domain gap and achieve better performance across different tasks and datasets.

  \begin{table*}[t!]
\setlength{\tabcolsep}{1pt}
 \centering 
{\footnotesize
\begin{tabular}{lc|ccc|c|ccc}   
\multicolumn{2}{c}{Method}    & $\mathcal{M}\rightarrow\mathcal{U}$ & $\mathcal{U}\rightarrow\mathcal{M}$ & $\mathcal{S}\rightarrow\mathcal{M}$ &Method & $\mathcal{M}\rightarrow\mathcal{U}$ & $\mathcal{U}\rightarrow\mathcal{M}$ & $\mathcal{S}\rightarrow\mathcal{M}$\\
\hline
\multicolumn{2}{c|}{GtA~\cite{sankaranarayanan2018generate}}& 92.8  $\pm$  0.9	&	90.8  $\pm$  1.3	&	92.4  $\pm$  0.9 &CDAN~\cite{long2018conditional} &93.9 &96.9& 88.5 \\
\multicolumn{2}{c|}{ADDA~\cite{tzeng2017adversarial}}& 89.4  $\pm$  0.2&90.1  $\pm$  0.8&76.0  $\pm$  1.8& ETD~\cite{li2020enhanced} & 96.4  $\pm$  0.3&96.3  $\pm$  0.1&97.9  $\pm$  0.4\\  
\multicolumn{2}{c|}{DWL~\cite{xiao2021dynamic}}&  97.3  &97.4& \textbf{98.1}& TDUDA~\cite{huang2022balancing}  & \textbf{97.8} & 97.1&97.3 \\  
\hline
\multicolumn{2}{c|}{RevGrad~\cite{ganin2014unsupervised}}&	 77.1  $\pm$  1.8 	&	73.0  $\pm$  2.0 	&	73.9  &JDDA~\cite{chen2019joint} & -& $97.0$ $\pm$0.2 & 93.1$\pm$0.2   \\ 
\multicolumn{2}{c|}{DRCN~\cite{ghifary2016deep}}&	 91.8  $\pm$  0.1 	&	73.7  $\pm$  0.4 	&	82.0  $\pm$  0.2 &  DMRL~\cite{wu2020dual}& 96.1 & \textbf{99.0} &96.2   \\
\hline
\multicolumn{2}{c|}{Source Only}&	 90.1$\pm$2.6	&	80.2$\pm$5.7	&	67.3$\pm$2.6   & Ours &    96.6  $\pm$  0.4	&	 98.3  $\pm$  0.3	& 96.6   $\pm$  0.9 \\
\end{tabular}}
\caption{Performance comparison for UDA tasks between MINIST, USPS, and SVHN datasets.}
\label{table:tabDA1}
 \end{table*}

Table~\ref{table:tabDA2} presents the comparison results for the UDA tasks of the Office-31 dataset. Upon examining the table, we observe that, on average, IMUDA achieves the best performance results among the compared methods. Furthermore, IMUDA outperforms other methods and leads to the best results in two of the tasks, while it is competitive in the remaining tasks.
The competitive performance of IMUDA can be attributed to several factors. Firstly, in the $\mathcal{D}\rightarrow \mathcal{W}$ and $\mathcal{W}\rightarrow \mathcal{D}$ tasks, the ``source only'' performance is  high, nearly 100\%. The high performance indicates that the initial domain gap between these two tasks is relatively small, meaning that the distributions of the source and target domains are already well matched prior to model adaptation. In such cases, inducing larger margins through our algorithm may not provide significant benefits since the interclass margins in the target domain are already relatively large and similar to those in the source domain. Consequently, there is no violation of margins prior to adaptation, limiting the potential improvement that can be obtained by increasing the margins further.
Although IMUDA's advantage in these tasks may be less pronounced due to the initially well-matched distributions, it is still able to deliver competitive performance. This suggests that IMUDA possesses robust adaptation capabilities and can effectively adapt the model to the target domain even when the domain gap is relatively small.
IMUDA's  performance in the Office-31 dataset demonstrates its effectiveness in bridging the domain gap and adapting the model to different target domains.

  \begin{table*}[t!]
  \setlength{\tabcolsep}{1pt}
 \centering 
{\footnotesize
\begin{tabular}{lc|cccccc|c}   
\multicolumn{2}{c}{Method}    & $\mathcal{A}\rightarrow\mathcal{W}$ & $\mathcal{D}\rightarrow\mathcal{W}$ & $\mathcal{W}\rightarrow\mathcal{D}$ &$\mathcal{A}\rightarrow\mathcal{D}$ &$\mathcal{D}\rightarrow\mathcal{A}$ &$\mathcal{W}\rightarrow\mathcal{A}$ & Average\\
\hline
\multicolumn{2}{c|}{Source Only~\cite{he2016deep}}&68.4  $\pm$  0.2 & 96.7  $\pm$  0.1&  99.3  $\pm$  0.1&  68.9  $\pm$  0.2 & 62.5  $\pm$  0.3&  60.7  $\pm$  0.3& 76.1  \\
\hline
\multicolumn{2}{c|}{GtA~\cite{sankaranarayanan2018generate}}&89.5  $\pm$  0.5& 97.9  $\pm$  0.3& 99.8  $\pm$  0.4& 87.7  $\pm$  0.5& 72.8  $\pm$  0.3& 71.4  $\pm$  0.4& 86.5 \\
\multicolumn{2}{c|}{DANN~\cite{ganin2016domain}}&   82.0  $\pm$  0.4& 96.9  $\pm$  0.2& 99.1  $\pm$  0.1& 79.7  $\pm$  0.4& 68.2  $\pm$  0.4 &67.4  $\pm$  0.5 &82.2 \\ 
\multicolumn{2}{c|}{ADDA~\cite{tzeng2017adversarial}}& 86.2  $\pm$  0.5 & 96.2  $\pm$  0.3&  98.4  $\pm$  0.3&  77.8  $\pm$  0.3 & 69.5  $\pm$  0.4&  68.9  $\pm$  0.5&82.8 \\
\multicolumn{2}{c|}{MADA~\cite{pei2018multi}}& 82.0  $\pm$  0.4&  96.9  $\pm$  0.2 & 99.1  $\pm$  0.1&  79.7  $\pm$  0.4 & 68.2  $\pm$  0.4&  67.4  $\pm$  0.5  &82.2 \\ 
\multicolumn{2}{c|}{CDAN~\cite{long2018conditional} }&93.1  $\pm$  0.2 &98.2  $\pm$  0.2& \textbf{\textbf{100}.0}  $\pm$  0.0& 89.8  $\pm$  0.3& 70.1 $\pm$  0.4& 68.0  $\pm$  0.4 &86.6\\  
\multicolumn{2}{c|}{DMRL~\cite{wu2020dual} }& 90.8$\pm$0.3&99.0$\pm$0.2&\textbf{100}.0$\pm$0.0&93.4$\pm$0.5&73.0$\pm$0.3 &   71.2$\pm$0.3&87.9 \\ 
\multicolumn{2}{c|}{DWL~\cite{xiao2021dynamic} }& 89.2& 99.2 & \textbf{100.0}& 91.2& 73.1& 69.8& 87.1 \\ \multicolumn{2}{c|}{ALSDA~\cite{mei2023automatic} }& 95.2& 99.2& \textbf{100.0} &95.8 &78.1 &77.5 &91.0 \\
\hline
\multicolumn{2}{c|}{DAN~\cite{long2015learning}}&  80.5 $\pm$ 0.4 &97.1 $\pm$ 0.2& 99.6 $\pm$ 0.1& 78.6 $\pm$ 0.2& 63.6 $\pm$ 0.3& 62.8 $\pm$ 0.2&80.4\\ %
\multicolumn{2}{c|}{DRCN~\cite{ghifary2016deep}}&	 72.6  $\pm$  0.3 	&	96.4  $\pm$  0.1	& 99.2  $\pm$  0.3 	& 67.1  $\pm$  0.3 	& 56.0  $\pm$  0.5 	&$72.6$ $\pm$  0.3 & 77.7	  \\ 
\multicolumn{2}{c|}{RevGrad~\cite{ganin2014unsupervised}} &82.0  $\pm$  0.4&96.9  $\pm$  0.2& 99.1  $\pm$  0.1& 79.7  $\pm$  0.4& 68.2  $\pm$  0.4& 67.4  $\pm$  0.5 &82.2\\
\multicolumn{2}{c|}{CADA-P~\cite{kurmi2019attending}}&83.4$\pm$0.2 &99.8$\pm$0.1  & \textbf{100}.0$\pm$0   &80.1$\pm$0.1 &59.8$\pm$0.2 &59.5$\pm$0.3 &80.4\\ 
\multicolumn{2}{c|}{JAN~\cite{long2017deep}}& 85.4  $\pm$  0.3& 97.4  $\pm$  0.2& 99.8  $\pm$  0.2& 84.7  $\pm$  0.3& 68.6  $\pm$  0.3 &70.0  $\pm$  0.4 &84.3 \\ 
\multicolumn{2}{c|}{JDDA~\cite{chen2019joint}}&82.6  $\pm$  0.4& 95.2  $\pm$  0.2& 99.7  $\pm$  0.0 &79.8  $\pm$  0.1 &57.4  $\pm$  0.0& 66.7  $\pm$  0.2 &80.2\\ 
\multicolumn{2}{c|}{ETD~\cite{li2020enhanced}}&92.1&\textbf{100}.0 & \textbf{\textbf{100}.0}&88.0&71.0&67.8  86.2&   86.5\\ 
\multicolumn{2}{c|}{MetaAlign~\cite{wei2021metaalign}}& 93.0 $\pm$ 0.5& 98.6 $\pm$ 0.0& \textbf{100} $\pm$ 0.0& 94.5 $\pm$ 0.3& 75.0 $\pm$ 0.3& 73.6 $\pm$ 0.0& 89.2\\
\multicolumn{2}{c|}{HCL~\cite{huang2021model}}&92.5 &98.2& \textbf{100.0}& 94.7 & 75.9 &77.7& 89.8   \\
\multicolumn{2}{c|}{FixBi~\cite{na2021fixbi}}&96.1$\pm$0.2& 99.3$\pm$0.2& \textbf{100.0}$\pm$0.0 &95.0$\pm$0.4& \textbf{78.7}$\pm$0.5 &\textbf{79.4}$\pm$0.3& 91.4 \\
\multicolumn{2}{c|}{CAF~\cite{xie2022collaborative}}&94.7$\pm$0.1& 98.7$\pm$0.2& \textbf{100.0}$\pm$0.0 &93.6$\pm$0.1&  72.7 $\pm$0.4 & 72.3$\pm$0.3& 88.4 \\\multicolumn{2}{c|}{TDUDA~\cite{huang2022balancing}}& 88.4 & 99.2 & \textbf{100.0} &89.1  & 71.8 &70.8  & 86.6 \\
\multicolumn{2}{c|}{PICSCS~\cite{li2023pseudo}}&93.6 &99.1 & \textbf{100.0}& 93.6& 77.1& 78.0& 90.2  \\\multicolumn{2}{c|}{COT~\cite{liu2023cot}}&96.5 &99.1& \textbf{100.0}& 96.1 &76.7& 77.4 &91.0 \\
 \hline
\multicolumn{2}{c|}{IMUDA}&	 \textbf{99.6}  $\pm$  0.2	&	98.1  $\pm$  0.2	&	99.6  $\pm$  0.1 & \textbf{99.0}  $\pm$  0.4   & 74.9  $\pm$  0.4   & 78.7  $\pm$  1.1 & \textbf{91.7}\\   
\end{tabular}}
\caption{ Performance comparison  for UDA tasks for  Office-31 dataset. }
\label{table:tabDA2}
 \end{table*}

The results for the ImageCLEF-DA dataset can be found in Table~\ref{table:tabDA3}. Upon examining the table, we can   see that IMUDA achieves a significant performance boost compared to previous UDA methods on this dataset. On average, IMUDA outperforms the next best method by approximately 7\%.
This   performance improvement can be attributed to the specific characteristics of the ImageCLEF-DA dataset. One notable aspect is that the dataset is fully balanced, meaning that it has an equal number of data points across both the domains and the classes. This balanced nature of the dataset plays a crucial role in the effectiveness of IMUDA. Matching the internal distributions with a Gaussian Mixture Model (GMM) distribution becomes more accurate in this scenario. Since the source dataset is balanced, the empirical source distribution used for domain alignment becomes more representative of the true source distribution in the ImageCLEF-DA dataset.
Moreover, the empirical interclass margin, which is an important factor in IMUDA, holds the same meaning across the domains due to the balanced data distribution. This consistency allows IMUDA to leverage the interclass margin effectively for adaptation, leading to improved performance.
From these observations, we can conclude that having a balanced training dataset in the source domain plays a significant role in boosting the performance of IMUDA on the ImageCLEF-DA dataset. It ensures that the empirical distributions and interclass margins are more accurately estimated and align well with the true underlying distributions in the dataset.
However, it is worth noting that one area for further improvement in our algorithm lies in replicating similar levels of performance improvement when the training dataset is imbalanced. While the balanced nature of the ImageCLEF-DA dataset benefits IMUDA, it is important to develop techniques that can achieve comparable performance gains even in the presence of imbalanced training datasets. Addressing this challenge would enhance the versatility and applicability of IMUDA in a broader range of real-world scenarios.

  \begin{table*}[t!]
  \setlength{\tabcolsep}{1pt}
 \centering 
{\footnotesize
\begin{tabular}{lc|ccccccc}   
\multicolumn{2}{c}{Method}    & $\mathcal{I}\rightarrow\mathcal{P}$ & $\mathcal{P}\rightarrow\mathcal{I}$ & $\mathcal{I}\rightarrow\mathcal{C}$ &$\mathcal{C}\rightarrow\mathcal{I}$ &$\mathcal{C}\rightarrow\mathcal{P}$ &$\mathcal{P}\rightarrow\mathcal{C}$& Average \\
\hline
\multicolumn{2}{c|}{Source Only~\cite{he2016deep}}& 74.8  $\pm$  0.3& 83.9  $\pm$  0.1& 91.5  $\pm$  0.3 &78.0  $\pm$  0.2 &65.5  $\pm$  0.3& 91.2  $\pm$  0.3 & 80.8 \\
\hline
\multicolumn{2}{c|}{DANN~\cite{ganin2016domain}}&   82.0  $\pm$  0.4& 96.9  $\pm$  0.2& 99.1 $\pm$  0.1& 79.7  $\pm$  0.4& 68.2  $\pm$  0.4 &67.4  $\pm$  0.5 &82.2 \\ 
\multicolumn{2}{c|}{MADA~\cite{pei2018multi}}& 75.0 $\pm$ 0.3& 87.9 $\pm$ 0.2& 96.0 $\pm$ 0.3& 88.8 $\pm$ 0.3& 75.2 $\pm$ 0.2& 92.2 $\pm$ 0.3 &85.9   \\
\multicolumn{2}{c|}{CDAN~\cite{long2018conditional} }&76.7 $\pm$ 0.3& 90.6 $\pm$ 0.3& 97.0 $\pm$ 0.4 &90.5 $\pm$ 0.4& 74.5 $\pm$ 0.3 &93.5 $\pm$ 0.4 & 87.1\\ 
\multicolumn{2}{c|}{DMRL~\cite{wu2020dual} }&77.3$\pm$0.4&90.7$\pm$0.3&97.4$\pm$0.3&91.8$\pm$0.3&76.0$\pm$0.5&94.8$\pm$0.3& 88.0 \\ 
\multicolumn{2}{c|}{DWL~\cite{xiao2021dynamic} }&
82.3 &94.8& 98.1& 92.8& 77.9& 97.2 &90.5\\
\hline
\multicolumn{2}{c|}{DAN~\cite{long2015learning}}& 74.5 $\pm$ 0.4 &82.2 $\pm$ 0.2 &92.8 $\pm$ 0.2& 86.3 $\pm$ 0.4& 69.2 $\pm$ 0.4& 89.8 $\pm$ 0.4&82.4\\ 
\multicolumn{2}{c|}{RevGrad~\cite{ganin2014unsupervised}} &75.0 $\pm$ 0.6 &86.0 $\pm$ 0.3& 96.2 $\pm$ 0.4 &87.0 $\pm$ 0.5& 74.3 $\pm$ 0.5& 91.5 $\pm$ 0.6&85.0 \\
\multicolumn{2}{c|}{JAN~\cite{long2017deep}}&  76.8 $\pm$ 0.4&  88.0 $\pm$ 0.2&  94.7 $\pm$ 0.2 & 89.5 $\pm$ 0.3&  74.2 $\pm$ 0.3 & 91.7 $\pm$ 0.3&85.8\\
\multicolumn{2}{c|}{CADA-P~\cite{kurmi2019attending}} &78.0&90.5 & 96.7&92.0 & 77.2&  95.5 & 88.3\\\multicolumn{2}{c|}{ETD~\cite{li2020enhanced}} &81.0 & 91.7 & 97.9  &93.3 & 79.5  &95.0&89.7\\
\multicolumn{2}{c|}{CGDM~\cite{du2021cross}}&  78.7 $\pm$ 0.2& 93.3 $\pm$ 0.1& 97.5 $\pm$ 0.3& 92.7 $\pm$ 0.2& 79.2 $\pm$ 0.1& 95.7 $\pm$ 0.2& 89.5\\
\multicolumn{2}{c|}{TDUDA~\cite{huang2022balancing}}& 80.3 & 93.2& 96.5& 90.5& 76.3&96.0 & 88.8\\
\multicolumn{2}{c|}{CAF~\cite{xie2022collaborative}}& 78.7 & 92.2& 98.0& 92.7& 77.8& 95.8& 89.2\\\multicolumn{2}{c|}{PICSCS~\cite{li2023pseudo}}& 81.7 & 94.8 & 96.8 & 95.8 &81.7  &96.5  &91.3\\
\hline
\multicolumn{2}{c|}{IMUDA}&	 		\textbf{89.5}  $\pm$  1.2	&	\textbf{99.8}  $\pm$  0.2 & \textbf{\textbf{100}}  $\pm$  0.0   & \textbf{99.9}  $\pm$  0.1 & \textbf{92.6}  $\pm$  0.9 & \textbf{99.8}  $\pm$  0.2  &\textbf{96.9}    \\
\end{tabular}}
\caption{ Performance comparison    for UDA tasks for  ImageCLEF-DA dataset. }
\label{table:tabDA3}
 \end{table*}

 The performance results for the single task of the VisDA2017 dataset can be found in Table~\ref{table:tabDA5}. Upon analyzing the table, we can see that IMUDA achieves a competitive performance on this dataset.
One key factor that contributes to the effectiveness of IMUDA on the VisDA2017 dataset is the large size of the dataset itself. The dataset contains a substantial number of samples, which in turn makes the empirical probability distribution a more accurate representation of the true underlying distribution. As a result, the empirical SWD loss employed by IMUDA can effectively enforce domain alignment, leading to improved performance.
Intuitively, we can understand that having larger training datasets plays a crucial role in boosting the performance of IMUDA on the VisDA2017 dataset. The increased sample size allows for a more comprehensive coverage of the underlying data distribution, enabling better alignment between the source and target domains. By leveraging the rich diversity and abundance of data, IMUDA can exploit the patterns and characteristics present in the dataset more effectively, leading to enhanced performance.
It is important to note that the relationship between dataset size and performance improvement is not solely dependent on the number of samples, but also on the quality and diversity of the data. A larger dataset   captures the intricate variations and complexities present in the real-world data, thereby facilitating more accurate domain alignment and adaptation.
In conclusion, the competitive performance of IMUDA on the VisDA2017 dataset can be attributed to the large size of the dataset.

Based on the results presented in Tables 1 to 4, it can be concluded that IMUDA, despite its simplicity, is a competitive UDA method when compared to existing methods. It either outperforms the existing methods or demonstrates performance that is comparable to the best-performing method in the respective UDA tasks.
It is important to note that due to the diverse nature of benchmark UDA tasks, no single UDA method can claim superiority over all other existing UDA methods across all major UDA tasks in the literature. The performance of each algorithm is highly dependent on various factors, including the specific task, dataset characteristics, hyper-parameter/parameter fine-tuning, and the choice of optimization techniques.
It is also worth mentioning that the performance of a given algorithm can vary to some extent by adjusting hyper-parameters/parameters and employing different optimization techniques. Therefore, algorithms with similar performance levels should be considered equally competitive.

Upon analyzing the tables, it can be observed that each algorithm leads to the best performance only on a subset of the UDA tasks. This highlights the importance of considering the specific circumstances and characteristics of each task when selecting an appropriate UDA method. The diversity of experiments conducted in this study has been instrumental in identifying the circumstances under which IMUDA is likely to yield better results.
In conclusion, IMUDA has demonstrated its competitiveness as a UDA method when compared to existing works, as evidenced by its superior performance or close performance proximity to the best-performing methods. The variability in performance across different UDA tasks emphasizes the need to carefully consider task-specific factors and select the most suitable algorithm accordingly.  These results have comparison nature but do not offer much insight about IMUDA. We offer additional analytic experiments 
for this purpose.

  \begin{table*}[t!]
  \setlength{\tabcolsep}{1pt}
 \centering 
{\footnotesize
\begin{tabular}{lc|cccccccccccc|c}   
\multicolumn{2}{c}{Method}    & Plane& Bike &Bus &Car& Horse &Knife& Motor& Person &Plant &Skateboard& Train& Truck & Average\\
\hline
\multicolumn{2}{c|}{Source Only}&70.6& 51.8& 55.8& 68.9 &77.9& 7.6 &93.3 &34.5& 81.1 &27.9 &88.6 &5.6 &55.3  \\
\hline
\multicolumn{2}{c|}{DANN \cite{ganin2016domain}}&81.9 &77.7 &82.8 &44.3 &81.2 &29.5 &65.1 &28.6& 51.9 &54.6 &82.8 &7.8& 57.4 \\
\multicolumn{2}{c|}{RevGrad \cite{ganin2014unsupervised}}& 75.9 &70.5 &65.3& 17.3 &72.8 &38.6 &58.0 &77.2 &72.5 &40.4 &70.4 &44.7 &58.6 \\
\multicolumn{2}{c|}{JAN \cite{long2017deep}}& 92.1&  66.4&  81.4 & 39.6 & 72.5 & 70.5 & 81.5&  70.5&  79.7 & 44.6 & 74.2 & 24.6 & 66.5 \\
\multicolumn{2}{c|}{GtA \cite{sankaranarayanan2018generate}}&  - &- &- &- &- &- &- &- &- &- &- &- & 77.1 \\
\multicolumn{2}{c|}{CDAN \cite{long2018conditional}}& 85.2&  66.9&  83.0&  50.8&  84.2& 74.9&  88.1&  74.5&  83.4&  76.0&  81.9 & 38.0&  73.7 \\
\multicolumn{2}{c|}{DMRL \cite{wu2020dual}}& - &- &- &- &- &- &- &- &- &- &- &- & 75.5\\
\multicolumn{2}{c|}{DAN \cite{long2015learning}}& 68.1 &15.4& 76.5& 87 &71.1& 48.9& 82.3 &51.5& 88.7 &33.2& 88.9 &42.2 &61.1 \\
\multicolumn{2}{c|}{DWL \cite{xiao2021dynamic}}&90.7& 80.2& 86.1& 67.6 &92.4 &81.5 &86.8 &78.0 &90.6 &57.1 &85.6 &28.7 &77.1  \\
\multicolumn{2}{c|}{CGDM \cite{du2021cross}}& 93.4 &82.7 &73.2 &68.4& 92.9 &94.5& 88.7& 82.1 &93.4& 82.5& 86.8& 49.2& 82.3 \\
\multicolumn{2}{c|}{ALSDA~\cite{mei2023automatic}}& 93.8& 72.8 &81.0 &49.0 &82.9 &90.5 &89.3 &80.8 &88.5 &86.6& 87.3& 43.9 &78.9\\
\multicolumn{2}{c|}{HCL \cite{huang2021model}}& 93.3 &85.4& 80.7& 68.5& 91.0& 88.1& 86.0 &78.6& 86.6 &88.8 &80.0 &74.7& 83.5 \\
\multicolumn{2}{c|}{BiFix \cite{na2021fixbi}}& 96.1 & 87.8  &90.5 &\textbf{90.3}& \textbf{96.8} &95.3 &92.8& \textbf{88.7}& \textbf{97.2}& 94.2& 90.9& 25.7& \textbf{87.2}\\
\multicolumn{2}{c|}{CAF~\cite{xie2022collaborative}}& 94.2  & 82.3 & 86.8 &68.9  & 87.3 & 93.3 & 88.4 &76.8  &92.9  & 69.7 & 83.5 &33.8&79.8\\\multicolumn{2}{c|}{COT~\cite{liu2023cot}}&96.9& \textbf{89.6} &84.2& 74.1 &96.4 &96.5& 88.6& 82.0 &96.0& 94.1 &85.1& 62.1& 87.1\\
 \hline
\multicolumn{2}{c|}{IMUDA}&	 \textbf{98.5}& 63.9& \textbf{92.8}& 74.9& 84.4& \textbf{98.8}& \textbf{93.9}& 86.1& 92.7& \textbf{95.5}& \textbf{94.2}&\textbf{45.3}& 85.1 \\   
\end{tabular}}
\caption{Performance    for the VisDA UDA task. }
\label{table:tabDA5}
 \end{table*}


\subsection{Analytic and Ablative  Analysis}

We conducted an empirical analysis of our algorithm to gain a deeper understanding of its impact on data representation. To assess the effect of the algorithm on aligning distributions in the embedding space, we focused on the $\mathcal{C}\rightarrow \mathcal{P}$ task of the ImageClef-DA dataset. Specifically, we utilized the testing split of data from both domains and projected them into the embedding space. To visualize the results in a more interpretable manner, we employed the UMAP (Uniform Manifold Approximation and Projection) visualization tool, which reduces the dimensionality of data representations to two for 2D visualization.

\begin{figure*}[t]
  \centering
    \begin{subfigure}[b]{0.24\textwidth}\includegraphics[width=\textwidth]{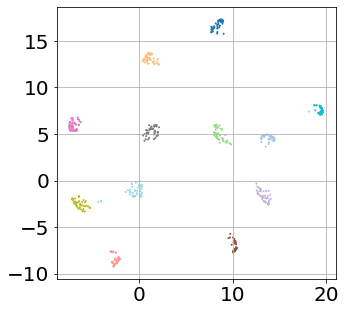}
        \caption{ }
        \label{fig11}
    \end{subfigure}
  \centering
    \begin{subfigure}[b]{0.24\textwidth}\includegraphics[width=\textwidth]{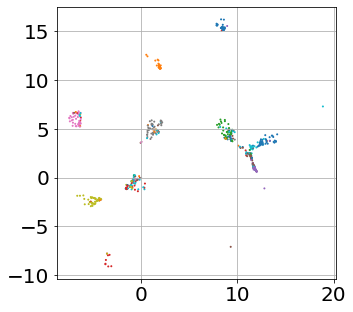}
        \caption{ }
        \label{fig12}
    \end{subfigure}
       \begin{subfigure}[b]{0.24\textwidth}\includegraphics[width=\textwidth]{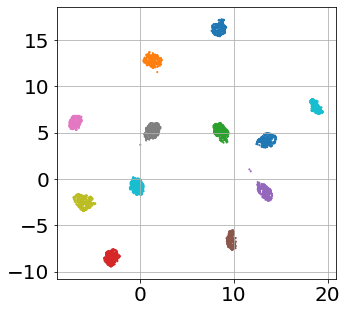}
           \centering
       \caption{ }
        \label{fig13}
    \end{subfigure}
      \centering
           \begin{subfigure}[b]{0.24\textwidth}\includegraphics[width=\textwidth]{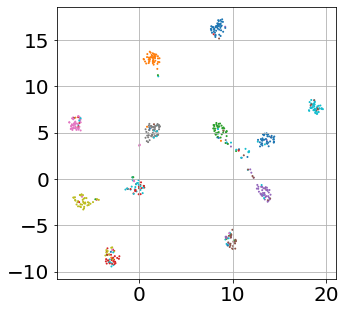}
           \centering
        \caption{ }
        \label{fig14}
    \end{subfigure}
     \caption{UMAP visualization for the representations of the dataset testing split for  the $\mathcal{C}\rightarrow \mathcal{P}$ task:  (a) the source domain (b) the target domain prior to adaptation, (c) samples drawn from the learned GMM, (d) the target domain after adaptation.  (Best viewed enlarged on screen and in color).  }\label{fig1T}
\end{figure*}

Figure~\ref{fig1T} displays the visualizations of the source domain testing split representations, samples drawn from the estimated GMM distribution, and the target domain testing split representations before and after performing unsupervised domain adaptation. Each point in the visualization represents an individual data point, and the colors correspond to the twelve different classes. By comparing Figures~\ref{fig11} and \ref{fig13}, we observe that high-confidence GMM samples align quite well with the source domain distribution for this specific task. This suggests that the GMM serves as a reliable parametric model for representing the source distribution. Furthermore, by comparing Figure~\ref{fig12} with Figure~\ref{fig11}, we notice that domain shift has caused overlapping clusters in the target domain, as previously depicted qualitatively in Figure~\ref{figMUDA:1}. However, Figure~\ref{fig14} demonstrates that the IMUDA algorithm effectively aligns the distribution of the target domain with the source domain distribution, mitigating the effects of domain shift. These empirical observations reinforce the underlying intuition behind our approach, as illustrated in Figure~\ref{figMUDA:1}.

By conducting this analysis and providing visual evidence of the alignment of distributions in the embedding space, we contribute to a better understanding of the efficacy of our IMUDA algorithm. The visualization results support the rationale behind the development of our algorithm and provide empirical validation of its ability to address the challenges posed by domain shift. This analysis enhances our confidence in the effectiveness of IMUDA as a means of achieving domain adaptation and improving the alignment of distributions between source and target domains. The results also matches with what the theoretical result predicts.

\begin{figure*}[t]
  \centering
    \begin{subfigure}[b]{0.24\textwidth}\includegraphics[width=\textwidth]{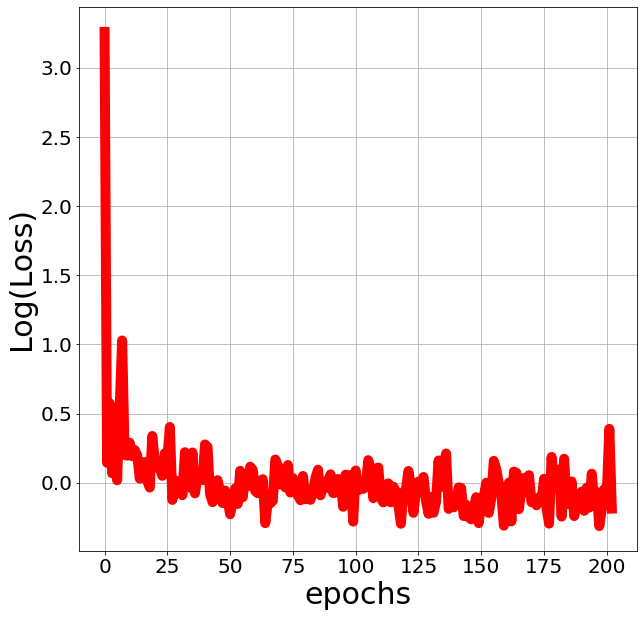}
        \caption{ }
        \label{fig21}
    \end{subfigure}
  \centering
    \begin{subfigure}[b]{0.22\textwidth}\includegraphics[width=\textwidth]{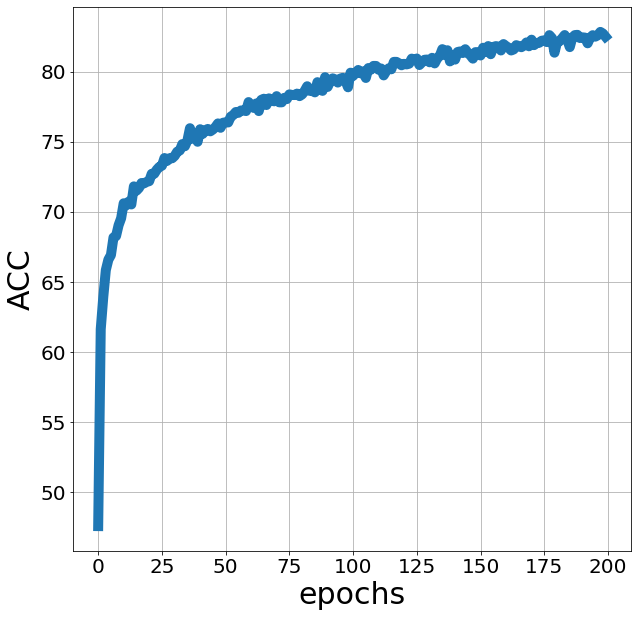}
        \caption{ }
        \label{fig22}
    \end{subfigure}
       \begin{subfigure}[b]{0.24\textwidth}\includegraphics[width=\textwidth]{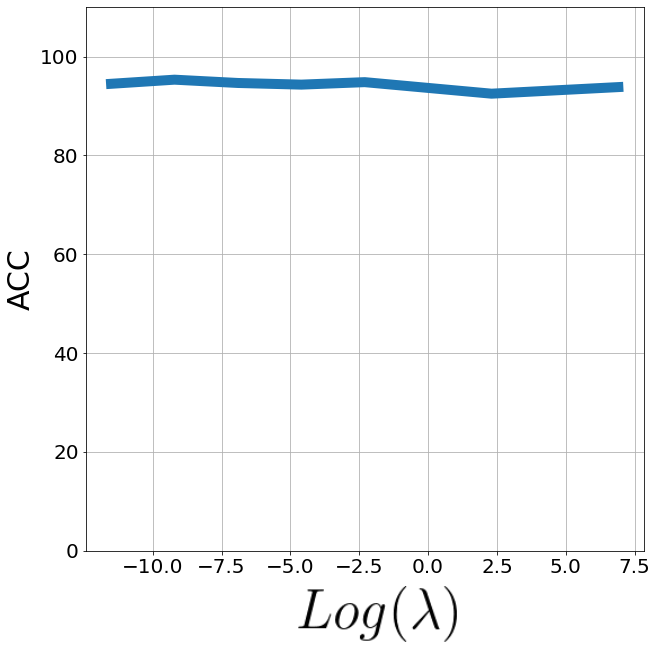}
           \centering
       \caption{ }
        \label{fig23}
    \end{subfigure}
      \centering
           \begin{subfigure}[b]{0.24\textwidth}\includegraphics[width=\textwidth]{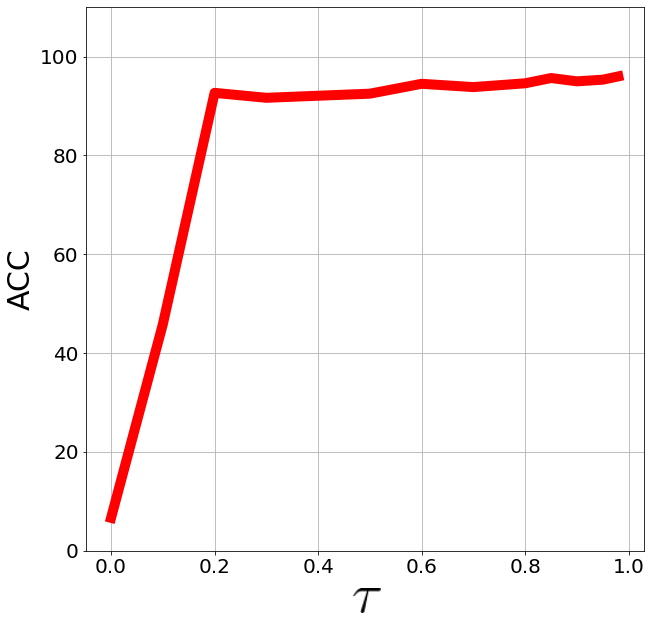}
           \centering
        \caption{ }
        \label{fig24}
    \end{subfigure}
     \caption{Empirical analysis based on  the VisDA task:  (a) loss function on the training split versus \#epochs and (b) learning curve for the testing split versus \#epochs; Effect of parameter values for the $\mathcal{C}\rightarrow \mathcal{P}$ task (c) performance versus the trade-off parameter $\lambda$ and (d) classification accuracy versus the confidence parameter $\tau$. (Best viewed enlarged on screen and in color).  }\label{fig2T}
\end{figure*}

We conducted additional investigations by examining the training loss function value and the classification performance on the testing split of the VisDA task in relation to the number of optimization epochs. The results are presented in Figures~\ref{fig21} and ~\ref{fig22}, respectively.
Figure~\ref{fig21} illustrates the training loss function value, which is calculated based on equation~\eqref{eq:smallmainPrMatch}. As the optimization epochs progress, we observe a decrease in the loss function value, indicating a reduction in the distance between the two empirical distributions. This trend aligns with our objective of achieving domain alignment through the IMUDA algorithm. By minimizing the loss function, we effectively bring the source and target domains closer together in the embedding space, facilitating better alignment and adaptation.
Figure~\ref{fig22} showcases the classification performance on the testing split of the VisDA task. As the optimization epochs increase, we observe an improvement in the classification performance on the target domain. This finding aligns with our expectation that enhancing domain alignment through the IMUDA algorithm would lead to improved model generalization on the target domain. The correlation between the decreasing loss function value and the increasing classification performance on the target domain further supports our reasoning that greater domain alignment results in better performance on the target domain data.  These experiments also serve as an empirical validation of Theorem 2.
 The observed relationship between the optimization objective loss function, domain alignment, and model performance on the target domain confirms that our algorithm successfully implements the desired domain alignment effect for UDA.

We conducted an analysis to explore the impact of the primary hyper-parameters of the IMUDA algorithm on performance. Figures~\ref{fig23} and ~\ref{fig24} depict the results of our experiments.
Figure~\ref{fig23} presents the classification accuracy across varying values of the hyper-parameter $\lambda$. It is evident that the UDA performance remains relatively constant as $\lambda$ varies. This outcome aligns with our expectations, as the empirical risk minimization (ERM) terms in equation~\eqref{eq:mainPrMatchICLR} are already minimized during the pretraining step, resulting in small ERM terms prior to UDA optimization. Consequently, the UDA optimization mainly focuses on minimizing the alignment loss terms, leading to consistent performance regardless of the specific value of $\lambda$.
Figure~\ref{fig24} showcases the classification accuracy for different values of the confidence hyper-parameter $\tau$. As anticipated from our previous discussions, we observe that larger values of $\tau$ yield improved performance. Conversely, very small values of $\tau$ can have a detrimental effect on performance, as low-confidence Gaussian mixture model (GMM) samples might behave as outliers, posing challenges for domain alignment compared to simple UDA methods based solely on distribution matching. Importantly, this experiment serves as an ablation study, confirming the critical role of using high-confidence samples for the IMUDA algorithm to enhance model generalization. The increasing performance with larger values of $\tau$, indicating a larger interclass margin in the source domain, demonstrates the effectiveness of our secondary mechanism in inducing larger margins. Notably, we observe that after a certain threshold (around $\tau\approx 0.2$), further increasing $\tau$ provides diminishing returns in terms of performance improvement. Although the performance still increases with a smaller slope for $\tau>0.2$, the rate of improvement diminishes. These results also confirm our theoretical analysis.

In addition, we conducted ablative experiments on the optimization terms in equation~\eqref{eq:smallmainPrMatch}. Specifically, we evaluated the impact of dropping the third and fourth terms to assess the contribution of each term to the overall performance using the Office-31 dataset. The results are presented in Table~\ref{table:abe}. Comparing the performances with IMUDA, we observe that both terms   contribute to achieving optimal performance, with the third term playing a more crucial role. This observation is expected since the third term directly enforces alignment between the target domain and the pseudo-dataset. On the other hand, the fourth term proves beneficial in inducing larger margins, complementing the primary alignment mechanism. Overall, the results emphasize that the fourth term serves as a complementary mechanism to enhance UDA results.
Through these   analyses, we elucidate the effects of the primary hyper-parameters, confirm the importance of using high-confidence samples, and demonstrate the contributions of individual optimization terms.

   \begin{table*}[t!]
     \setlength{\tabcolsep}{1pt}
 \centering 
{\footnotesize
\begin{tabular}{lc|cccccc|c}   
\multicolumn{2}{c}{Method}    & $\mathcal{A}\rightarrow\mathcal{W}$ & $\mathcal{D}\rightarrow\mathcal{W}$ & $\mathcal{W}\rightarrow\mathcal{D}$ &$\mathcal{A}\rightarrow\mathcal{D}$ &$\mathcal{D}\rightarrow\mathcal{A}$ &$\mathcal{W}\rightarrow\mathcal{A}$ & Average\\
\hline
\multicolumn{2}{c|}{$3^{rd}$ Term}&	  70.2 $\pm$ 0.9 & 91.3 $\pm$ 1.9 & 96.8 $\pm$    0.5&77.4 $\pm$ 0.4 & 58.1$\pm$0.3  & 57.7 $\pm$ 0.4 &  75.3\\
\multicolumn{2}{c|}{$4^{th}$ Term}&	  99.3 $\pm$ 0.3 & 95.1 $\pm$ 0.5 & 98.9 $\pm$    0.1& 99.0$\pm$ 0.1 &77.0 $\pm$0.5  & 77.8 $\pm$0.1  & 91.2 \\
\multicolumn{2}{c|}{IMUDA}&	 99.6  $\pm$  0.2	&	98.1  $\pm$  0.2	&	99.6  $\pm$  0.1 & 99.0  $\pm$  0.4   & 74.9  $\pm$  0.4   & 78.7  $\pm$  1.1 & 91.7\\   
\end{tabular}}
\caption{Ablative Experiments on Optimization Terms Using the  Office-31 dataset. }
\label{table:abe}
 \end{table*}

 \section{Conclusions  }
We   introduced a new unsupervised domain adaptation method to mitigate the influence of domain shift on model generalization. Our idea is based on expanding the distances between clusters of different classes within an embedding space. We represent this embedding space as the output-space of a deep neural encoder's final layer. By increasing the interclass distances, we enhance the model's ability to generalize in the target domain. We estimate the internal distribution of the source domain data and using the obtained knowledge, push the data representations   away from the class boundaries within the embedding space to increase robustness with respect to domain shift. To estimate the internal distribution, we employ a parametric Gaussian mixture model (GMM). Subsequently, we generate confident labeled random samples from the GMM and use them to create larger interclass distances.
Our theoretical analysis provides a justification to explain why algorithm can alleviate the impact of domain shift.
Our empirical results demonstrate the effectiveness of our approach, showing competitive performance when compared to existing unsupervised domain adaptation methods. Furthermore, we conclude that integrating additional mechanisms is crucial for improving domain alignment in unsupervised domain adaptation, especially given the current performance levels of such algorithms.

Despite strengths, our algorithm has its own limitations. The first major limitation is that we consider that a multimodal distribution is formed in the embedding space, where we know that each class is represented by a single mode. While we observed the validity of this assumption in the datasets and models that we used in our experiments,  there is no guarantee that this assumption will be valid in all models and datasets. If this assumption is not valid, then using our approach will not lead to having a good estimate for the internal distribution. A reasonable follow-up work is to address this weakness by proposing a stronger method to estimate the internal representation that relaxes our assumption.
A second limitation of our work is that, as we observed in the experiments, our method leads to a better performance when the datasets are balanced. Another direction for future research is addressing the scenario where the source dataset exhibits class imbalance. The third limitation is that the upperbound that we offer in our theoretical is not guaranteed to be tight. A suitable theoretical future work is to offer an estimate about the tightness of the upperbound given the source and the target domain data. Such an estimate can help to check how effective our approach can be given a source and a target domain before attempting using the full pipleine of our method. The fourth limitation of work is that we consider that the source domain samples are readily accessible during UDA but in some applications, the source sample are either unavailable or cannot be shared due to privacy concerns. In those situations, our method is not applicable and a new solution should be proposed. Another limitation that we have is that the pre-adaptation and the post-adaptation steps should be performed sequentially. An improvement for future exploration is to merge these two stages. Finally, we are considering a UDA setting where both domains share exactly the same classes. In practical settings, two domains may share only a subset of classes. Our approach is not directly applicable in those scenarios, i.e., partial domain adaptation, and extending to address partial domain adaptation is another future research possibility for our work.

\paragraph{\textbf{Data Availability}} Datasets are publicly available.

 \paragraph{\textbf{Conflict of Interest}} The author declares no known competing financial interests.

 {    
    \small
    \bibliographystyle{plain}
    \bibliography{main}

\begin{thebibliography}{10}

\bibitem{bhushan2018deepjdot}
Bharath Bhushan~Damodaran, Benjamin Kellenberger, R{\'e}mi Flamary, Devis Tuia,
  and Nicolas Courty.
\newblock Deepjdot: Deep joint distribution optimal transport for unsupervised
  domain adaptation.
\newblock In {\em ECCV}, pages 447--463, 2018.

\bibitem{damodaran2018deepjdot}
Bharath Bhushan~Damodaran, Benjamin Kellenberger, R{\'e}mi Flamary, Devis Tuia,
  and Nicolas Courty.
\newblock Deepjdot: Deep joint distribution optimal transport for unsupervised
  domain adaptation.
\newblock In {\em Proceedings of ECCV}, pages 447--463, 2018.

\bibitem{bonnotte2013unidimensional}
N.~Bonnotte.
\newblock {\em Unidimensional and evolution methods for optimal
  transportation}.
\newblock PhD thesis, Paris 11, 2013.

\bibitem{cao2019theoretical}
Tianshi Cao, Marc~T Law, and Sanja Fidler.
\newblock A theoretical analysis of the number of shots in few-shot learning.
\newblock In {\em International Conference on Learning Representations}, 2019.

\bibitem{chen2019joint}
Chao Chen, Zhihong Chen, Boyuan Jiang, and Xinyu Jin.
\newblock Joint domain alignment and discriminative feature learning for
  unsupervised deep domain adaptation.
\newblock In {\em Proceedings of the AAAI}, pages 3296--3303, 2019.

\bibitem{chen2020homm}
Chao Chen, Zhihang Fu, Zhihong Chen, Sheng Jin, Zhaowei Cheng, Xinyu Jin, and
  Xian-Sheng Hua.
\newblock Homm: Higher-order moment matching for unsupervised domain
  adaptation.
\newblock In {\em Proceedings of the AAAI conference on artificial
  intelligence}, volume~34, pages 3422--3429, 2020.

\bibitem{chen2023activate}
Chaoqi Chen, Luyao Tang, Leitian Tao, Hong-Yu Zhou, Yue Huang, Xiaoguang Han,
  and Yizhou Yu.
\newblock Activate and reject: Towards safe domain generalization under
  category shift.
\newblock In {\em Proceedings of the IEEE/CVF International Conference on
  Computer Vision}, pages 11552--11563, 2023.

\bibitem{chen2019progressive}
Chaoqi Chen, Weiping Xie, Wenbing Huang, Yu~Rong, Xinghao Ding, Yue Huang,
  Tingyang Xu, and Junzhou Huang.
\newblock Progressive feature alignment for unsupervised domain adaptation.
\newblock In {\em Proceedings of CVPR}, pages 627--636, 2019.

\bibitem{courty2017optimal}
Nicolas Courty, R{\'e}mi Flamary, Devis Tuia, and Alain Rakotomamonjy.
\newblock Optimal transport for domain adaptation.
\newblock {\em IEEE Transactions on Pattern Analysis and Machine Intelligence},
  39(9):1853--1865, 2016.

\bibitem{dhouib2020margin}
Sofien Dhouib, Ievgen Redko, and Carole Lartizien.
\newblock Margin-aware adversarial domain adaptation with optimal transport.
\newblock In {\em International Conference on Machine Learning}, pages
  2514--2524. PMLR, 2020.

\bibitem{du2021cross}
Zhekai Du, Jingjing Li, Hongzu Su, Lei Zhu, and Ke~Lu.
\newblock Cross-domain gradient discrepancy minimization for unsupervised
  domain adaptation.
\newblock In {\em Proceedings of the IEEE/CVF Conference on Computer Vision and
  Pattern Recognition}, pages 3937--3946, 2021.

\bibitem{el2022hierarchical}
Mourad El~Hamri, Younes Bennani, and Issam Falih.
\newblock Hierarchical optimal transport for unsupervised domain adaptation.
\newblock {\em Machine Learning}, 111(11):4159--4182, 2022.

\bibitem{gabourie2019learning}
Alexander~J Gabourie, Mohammad Rostami, Philip~E Pope, Soheil Kolouri, and
  Kuyngnam Kim.
\newblock Learning a domain-invariant embedding for unsupervised domain
  adaptation using class-conditioned distribution alignment.
\newblock In {\em 2019 57th Annual Allerton Conference on Communication,
  Control, and Computing (Allerton)}, pages 352--359. IEEE, 2019.

\bibitem{ganin2016domain}
Y.~Ganin, E.~Ustinova, H.~Ajakan, P.~Germain, H.~Larochelle, F.~Laviolette,
  M.~Marchand, and V.~Lempitsky.
\newblock Domain-adversarial training of neural networks.
\newblock {\em The Journal of Machine Learning Research}, 17(1):2096--2030,
  2016.

\bibitem{ganin2015unsupervised}
Yaroslav Ganin and Victor Lempitsky.
\newblock Unsupervised domain adaptation by backpropagation.
\newblock In {\em International conference on machine learning}, pages
  1180--1189. PMLR, 2015.

\bibitem{ganin2014unsupervised}
Yaroslav Ganin and Victor Lempitsky.
\newblock Unsupervised domain adaptation by backpropagation.
\newblock In {\em Proceedings of Proceedings of ICML}, pages 1180--1189, 2015.

\bibitem{ghifary2016deep}
Muhammad Ghifary, W~Bastiaan Kleijn, Mengjie Zhang, David Balduzzi, and Wen Li.
\newblock Deep reconstruction-classification networks for unsupervised domain
  adaptation.
\newblock In {\em Proceedings of ECCV}, pages 597--613. Springer, 2016.

\bibitem{he2016deep}
Kaiming He, Xiangyu Zhang, Shaoqing Ren, and Jian Sun.
\newblock Deep residual learning for image recognition.
\newblock In {\em Proceedings of CVPR}, pages 770--778, 2016.

\bibitem{hoffman2018cycada}
Judy Hoffman, Eric Tzeng, Taesung Park, Jun-Yan Zhu, Phillip Isola, Kate
  Saenko, Alexei Efros, and Trevor Darrell.
\newblock Cycada: Cycle-consistent adversarial domain adaptation.
\newblock In {\em Proceedings of ICML}, pages 1989--1998. PMLR, 2018.

\bibitem{huang2021model}
Jiaxing Huang, Dayan Guan, Aoran Xiao, and Shijian Lu.
\newblock Model adaptation: Historical contrastive learning for unsupervised
  domain adaptation without source data.
\newblock {\em Advances in Neural Information Processing Systems}, 34, 2021.

\bibitem{huang2022balancing}
Jingke Huang, Ni~Xiao, and Lei Zhang.
\newblock Balancing transferability and discriminability for unsupervised
  domain adaptation.
\newblock {\em IEEE Transactions on Neural Networks and Learning Systems},
  2022.

\bibitem{jian2023unsupervised}
Dayuan Jian and Mohammad Rostami.
\newblock Unsupervised domain adaptation for training event-based networks
  using contrastive learning and uncorrelated conditioning.
\newblock {\em arXiv preprint arXiv:2303.12424}, 2023.

\bibitem{kim2019unsupervised}
Minyoung Kim, Pritish Sahu, Behnam Gholami, and Vladimir Pavlovic.
\newblock Unsupervised visual domain adaptation: A deep max-margin gaussian
  process approach.
\newblock In {\em Proceedings of CVPR}, pages 4380--4390, 2019.

\bibitem{kim2022did}
Tae~Soo Kim, Geonwoon Jang, Sanghyup Lee, and Thijs Kooi.
\newblock Did you get what you paid for? rethinking annotation cost of deep
  learning based computer aided detection in chest radiographs.
\newblock In {\em Medical Image Computing and Computer Assisted
  Intervention--MICCAI 2022: 25th International Conference, Singapore,
  September 18--22, 2022, Proceedings, Part III}, pages 261--270. Springer,
  2022.

\bibitem{kundu2020universal}
Jogendra~Nath Kundu, Naveen Venkat, R~Venkatesh Babu, et~al.
\newblock Universal source-free domain adaptation.
\newblock In {\em Proceedings of the IEEE/CVF Conference on Computer Vision and
  Pattern Recognition}, pages 4544--4553, 2020.

\bibitem{kurmi2019attending}
Vinod~Kumar Kurmi, Shanu Kumar, and Vinay~P Namboodiri.
\newblock Attending to discriminative certainty for domain adaptation.
\newblock In {\em CVPR}, pages 491--500, 2019.

\bibitem{lee2019sliced}
Chen-Yu Lee, Tanmay Batra, Mohammad~Haris Baig, and Daniel Ulbricht.
\newblock Sliced wasserstein discrepancy for unsupervised domain adaptation.
\newblock In {\em Proceedings of CVPR}, pages 10285--10295, 2019.

\bibitem{li2020sequential}
Da~Li, Yongxin Yang, Yi-Zhe Song, and Timothy Hospedales.
\newblock Sequential learning for domain generalization.
\newblock In {\em European Conference on Computer Vision}, pages 603--619.
  Springer, 2020.

\bibitem{li2023pseudo}
Lei Li, Jun Yang, Yulin Ma, and Xuefeng Kong.
\newblock Pseudo-labeling integrating centers and samples with consistent
  selection mechanism for unsupervised domain adaptation.
\newblock {\em Information Sciences}, 628:50--69, 2023.

\bibitem{li2020enhanced}
Mengxue Li, Yi-Ming Zhai, You-Wei Luo, Peng-Fei Ge, and Chuan-Xian Ren.
\newblock Enhanced transport distance for unsupervised domain adaptation.
\newblock In {\em Proceedings of the IEEE/CVF Conference on Computer Vision and
  Pattern Recognition}, pages 13936--13944, 2020.

\bibitem{li2020model}
Rui Li, Qianfen Jiao, Wenming Cao, Hau-San Wong, and Si~Wu.
\newblock Model adaptation: Unsupervised domain adaptation without source data.
\newblock In {\em Proceedings of the IEEE/CVF Conference on Computer Vision and
  Pattern Recognition}, pages 9641--9650, 2020.

\bibitem{li2018adaptive}
Yanghao Li, Naiyan Wang, Jianping Shi, Xiaodi Hou, and Jiaying Liu.
\newblock Adaptive batch normalization for practical domain adaptation.
\newblock {\em Pattern Recognition}, 80:109--117, 2018.

\bibitem{li2020intelligent}
Yibin Li, Yan Song, Lei Jia, Shengyao Gao, Qiqiang Li, and Meikang Qiu.
\newblock Intelligent fault diagnosis by fusing domain adversarial training and
  maximum mean discrepancy via ensemble learning.
\newblock {\em IEEE Transactions on Industrial Informatics}, 17(4):2833--2841,
  2020.

\bibitem{liu2023cot}
Yang Liu, Zhipeng Zhou, and Baigui Sun.
\newblock Cot: Unsupervised domain adaptation with clustering and optimal
  transport.
\newblock In {\em Proceedings of the IEEE/CVF Conference on Computer Vision and
  Pattern Recognition}, pages 19998--20007, 2023.

\bibitem{long2015learning}
Mingsheng Long, Yue Cao, Jianmin Wang, and Michael Jordan.
\newblock Learning transferable features with deep adaptation networks.
\newblock In {\em Proceedings of Proceedings of ICML}, pages 97--105, 2015.

\bibitem{long2018conditional}
Mingsheng Long, Zhangjie Cao, Jianmin Wang, and Michael~I Jordan.
\newblock Conditional adversarial domain adaptation.
\newblock In {\em Proceedings of NeurIPS}, pages 1640--1650, 2018.

\bibitem{long2017deep}
Mingsheng Long, Han Zhu, Jianmin Wang, and Michael~I Jordan.
\newblock Deep transfer learning with joint adaptation networks.
\newblock In {\em Proceedings of the 34th Proceedings of ICML-Volume 70}, pages
  2208--2217. JMLR. org, 2017.

\bibitem{mei2023automatic}
Zhen Mei, Peng Ye, Hancheng Ye, Baopu Li, Jinyang Guo, Tao Chen, and Wanli
  Ouyang.
\newblock Automatic loss function search for adversarial unsupervised domain
  adaptation.
\newblock {\em IEEE Transactions on Circuits and Systems for Video Technology},
  2023.

\bibitem{moon1996expectation}
Todd~K Moon.
\newblock The expectation-maximization algorithm.
\newblock {\em IEEE Signal processing magazine}, 13(6):47--60, 1996.

\bibitem{motiian2017unified}
Saeid Motiian, Marco Piccirilli, Donald~A Adjeroh, and Gianfranco Doretto.
\newblock Unified deep supervised domain adaptation and generalization.
\newblock In {\em Proceedings of CVPR}, pages 5715--5725, 2017.

\bibitem{na2021fixbi}
Jaemin Na, Heechul Jung, Hyung~Jin Chang, and Wonjun Hwang.
\newblock Fixbi: Bridging domain spaces for unsupervised domain adaptation.
\newblock In {\em Proceedings of the IEEE/CVF Conference on Computer Vision and
  Pattern Recognition}, pages 1094--1103, 2021.

\bibitem{neal2003slice}
R.~Neal.
\newblock Slice sampling.
\newblock {\em Annals of statistics}, pages 705--741, 2003.

\bibitem{oza2023unsupervised}
Poojan Oza, Vishwanath~A Sindagi, Vibashan~Vishnukumar Sharmini, and Vishal~M
  Patel.
\newblock Unsupervised domain adaptation of object detectors: A survey.
\newblock {\em IEEE Transactions on Pattern Analysis and Machine Intelligence},
  2023.

\bibitem{pan2019transferrable}
Yingwei Pan, Ting Yao, Yehao Li, Yu~Wang, Chong-Wah Ngo, and Tao Mei.
\newblock Transferrable prototypical networks for unsupervised domain
  adaptation.
\newblock In {\em Proceedings of CVPR}, pages 2239--2247, 2019.

\bibitem{pei2018multi}
Zhongyi Pei, Zhangjie Cao, Mingsheng Long, and Jianmin Wang.
\newblock Multi-adversarial domain adaptation.
\newblock In {\em Proceedings Thirty-Second AAAI}, pages 3934--3941, 2018.

\bibitem{redko2017theoretical}
A.~Redko, I.and~Habrard and M.~Sebban.
\newblock Theoretical analysis of domain adaptation with optimal transport.
\newblock In {\em Joint European Conference on Machine Learning and Knowledge
  Discovery in Databases}, pages 737--753. Springer, 2017.

\bibitem{rostami2021lifelong}
Mohammad Rostami.
\newblock Lifelong domain adaptation via consolidated internal distribution.
\newblock {\em Advances in Neural Information Processing Systems},
  34:11172--11183, 2021.

\bibitem{rostami2022increasing}
Mohammad Rostami.
\newblock Increasing model generalizability for unsupervised visual domain
  adaptation.
\newblock In {\em Conference on Lifelong Learning Agents}, pages 281--293.
  PMLR, 2022.

\bibitem{rostami2021domain}
Mohammad Rostami and Aram Galstyan.
\newblock Domain adaptation for sentiment analysis using increased intraclass
  separation.
\newblock In {\em EMNLP}, 2021.

\bibitem{rostami2021cognitively}
Mohammad Rostami and Aram Galstyan.
\newblock Cognitively inspired learning of incremental drifting concepts.
\newblock In {\em International Joint Conference on Artificial Intelligence},
  2023.

\bibitem{rostami2023concpt}
Mohammad Rostami and Aram Galstyan.
\newblock Overcoming concept shift in domain-aware settings through
  consolidated internal distributions.
\newblock In {\em Thirty-Seventh AAAI Conference on Artificial Intelligence},
  2023.

\bibitem{rostami2022transfer}
Mohammad Rostami, Hangfeng He, Muhao Chen, and Dan Roth.
\newblock Transfer learning via representation learning.
\newblock In {\em Federated and Transfer Learning}, pages 233--257. Springer,
  2022.

\bibitem{rostami2018crowdsourcing}
Mohammad Rostami, David Huber, and Tsai-Ching Lu.
\newblock A crowdsourcing triage algorithm for geopolitical event forecasting.
\newblock In {\em Proceedings of the 12th ACM Conference on Recommender
  Systems}, pages 377--381. ACM, 2018.

\bibitem{rostami2019deep}
Mohammad Rostami, Soheil Kolouri, Eric Eaton, and Kyungnam Kim.
\newblock Deep transfer learning for few-shot sar image classification.
\newblock {\em Remote Sensing}, 11(11):1374, 2019.

\bibitem{rostami2019sar}
Mohammad Rostami, Soheil Kolouri, Eric Eaton, and Kyungnam Kim.
\newblock Sar image classification using few-shot cross-domain transfer
  learning.
\newblock In {\em Proceedings of the IEEE/CVF Conference on Computer Vision and
  Pattern Recognition Workshops}, pages 0--0, 2019.

\bibitem{rostami2021detection}
Mohammad Rostami, Leonidas Spinoulas, Mohamed Hussein, Joe Mathai, and Wael
  Abd-Almageed.
\newblock Detection and continual learning of novel face presentation attacks.
\newblock In {\em Proceedings of the IEEE/CVF International Conference on
  Computer Vision}, pages 14851--14860, 2021.

\bibitem{roweis1998algorithms}
Sam~T Roweis.
\newblock Em algorithms for pca and spca.
\newblock In {\em Proceedings of NeurIPS}, pages 626--632, 1998.

\bibitem{sankaranarayanan2018generate}
Swami Sankaranarayanan, Yogesh Balaji, Carlos~D Castillo, and Rama Chellappa.
\newblock Generate to adapt: Aligning domains using generative adversarial
  networks.
\newblock In {\em Proceedings of CVPR}, pages 8503--8512, 2018.

\bibitem{saporta2022multi}
Antoine Saporta, Arthur Douillard, Tuan-Hung Vu, Patrick P{\'e}rez, and
  Matthieu Cord.
\newblock Multi-head distillation for continual unsupervised domain adaptation
  in semantic segmentation.
\newblock In {\em Proceedings of the IEEE/CVF Conference on Computer Vision and
  Pattern Recognition}, pages 3751--3760, 2022.

\bibitem{shalev2014understanding}
Shai Shalev-Shwartz and Shai Ben-David.
\newblock {\em Understanding machine learning: From theory to algorithms}.
\newblock Cambridge university press, 2014.

\bibitem{stan2021unsupervised}
Serban Stan and Mohammad Rostami.
\newblock Unsupervised model adaptation for continual semantic segmentation.
\newblock In {\em Proceedings of the AAAI Conference on Artificial
  Intelligence}, volume~35, pages 2593--2601, 2021.

\bibitem{stan2021privacy}
Serban Stan and Mohammad Rostami.
\newblock Domain adaptation for the segmentation of confidential medical
  images.
\newblock In {\em British Machine Vision Conference}, 2022.

\bibitem{stan2022secure}
Serban Stan and Mohammad Rostami.
\newblock Secure domain adaptation with multiple sources.
\newblock {\em Transactions on Machine Learning Research}, 2022.

\bibitem{stan2022unsupervised}
Serban Stan and Mohammad Rostami.
\newblock Unsupervised model adaptation for source-free segmentation of medical
  images.
\newblock {\em arXiv preprint arXiv:2211.00807}, 2022.

\bibitem{stan2023preserving}
Serban Stan and Mohammad Rostami.
\newblock Preserving fairness in ai under domain shift.
\newblock {\em arXiv preprint arXiv:2301.12369}, 2023.

\bibitem{sun2016deep}
Baochen Sun and Kate Saenko.
\newblock Deep coral: Correlation alignment for deep domain adaptation.
\newblock In {\em Proceedings of ECCV}, pages 443--450. Springer, 2016.

\bibitem{tzeng2017adversarial}
Eric Tzeng, Judy Hoffman, Kate Saenko, and Trevor Darrell.
\newblock Adversarial discriminative domain adaptation.
\newblock In {\em Proceedings of CVPR}, pages 7167--7176, 2017.

\bibitem{wei2021metaalign}
Guoqiang Wei, Cuiling Lan, Wenjun Zeng, and Zhibo Chen.
\newblock Metaalign: Coordinating domain alignment and classification for
  unsupervised domain adaptation.
\newblock In {\em Proceedings of the IEEE/CVF Conference on Computer Vision and
  Pattern Recognition}, pages 16643--16653, 2021.

\bibitem{westfechtel2024gradual}
Thomas Westfechtel, Hao-Wei Yeh, Dexuan Zhang, and Tatsuya Harada.
\newblock Gradual source domain expansion for unsupervised domain adaptation.
\newblock In {\em Proceedings of the IEEE/CVF Winter Conference on Applications
  of Computer Vision}, pages 1946--1955, 2024.

\bibitem{wu2023unsupervised}
Mengxi Wu and Mohammad Rostami.
\newblock Unsupervised domain adaptation for graph-structured data using
  class-conditional distribution alignment.
\newblock {\em arXiv preprint arXiv:2301.12361}, 2023.

\bibitem{wu2020dual}
Yuan Wu, Diana Inkpen, and Ahmed El-Roby.
\newblock Dual mixup regularized learning for adversarial domain adaptation.
\newblock In {\em Proceedings of ECCV}, pages 540--555. Springer, 2020.

\bibitem{xiao2021dynamic}
Ni~Xiao and Lei Zhang.
\newblock Dynamic weighted learning for unsupervised domain adaptation.
\newblock In {\em Proceedings of the IEEE/CVF Conference on Computer Vision and
  Pattern Recognition}, pages 15242--15251, 2021.

\bibitem{xie2022collaborative}
Binhui Xie, Shuang Li, Fangrui Lv, Chi~Harold Liu, Guoren Wang, and Dapeng Wu.
\newblock A collaborative alignment framework of transferable knowledge
  extraction for unsupervised domain adaptation.
\newblock {\em IEEE Transactions on Knowledge and Data Engineering}, 2022.

\bibitem{xu2020reliable}
Renjun Xu, Pelen Liu, Liyan Wang, Chao Chen, and Jindong Wang.
\newblock Reliable weighted optimal transport for unsupervised domain
  adaptation.
\newblock In {\em Proceedings of the IEEE/CVF Conference on Computer Vision and
  Pattern Recognition}, pages 4394--4403, 2020.

\bibitem{zellinger2016central}
Werner Zellinger, Thomas Grubinger, Edwin Lughofer, Thomas Natschl{\"a}ger, and
  Susanne Saminger-Platz.
\newblock Central moment discrepancy (cmd) for domain-invariant representation
  learning.
\newblock In {\em ICLR}, 2017.

\bibitem{zhang2020unsupervised}
Dejiao Zhang, Ramesh Nallapati, Henghui Zhu, Feng Nan, Cicero dos Santos,
  Kathleen McKeown, and Bing Xiang.
\newblock Unsupervised domain adaptation for cross-lingual text labeling.
\newblock In {\em Proceedings of the 2020 Conference on Empirical Methods in
  Natural Language Processing: Findings}, pages 3527--3536, 2020.

\bibitem{zhang2021adaptive}
Marvin Zhang, Henrik Marklund, Nikita Dhawan, Abhishek Gupta, Sergey Levine,
  and Chelsea Finn.
\newblock Adaptive risk minimization: Learning to adapt to domain shift.
\newblock {\em Advances in Neural Information Processing Systems},
  34:23664--23678, 2021.

\bibitem{zhang2024category}
Siyu Zhang, Lei Su, Jiefei Gu, Ke~Li, Weitian Wu, and Michael Pecht.
\newblock Category-level selective dual-adversarial network using
  significance-augmented unsupervised domain adaptation for surface defect
  detection.
\newblock {\em Expert Systems with Applications}, 238:121879, 2024.

\bibitem{zhang2020discriminative}
Wen Zhang and Dongrui Wu.
\newblock Discriminative joint probability maximum mean discrepancy (djp-mmd)
  for domain adaptation.
\newblock In {\em 2020 international joint conference on neural networks
  (IJCNN)}, pages 1--8. IEEE, 2020.

\end{thebibliography}
}

\appendix

\section{Definition and Computation of Sliced Wasserstien Distance}

 SWD is defined by employing the idea of slicing on the  Wasserstein distance (WD) when the two distributions are high dimensional. WD distance is defined in terms of the following optimization problem:  
\begin{equation}
W(p_{\mathcal{S}},p_{\mathcal{T}})=\text{inf}_{\gamma\in \Gamma(p_{\mathcal{S}},p_{\mathcal{T}})} \int_{{X}\times {Y}} c(x,y)d\gamma(x,y),
\label{eq:kantorovich}
\end{equation}
where the set $\Gamma(p_{\mathcal{S}}, p_{\mathcal{T}})$ represents the collection of joint distributions $p_{{\mathcal{S}},{\mathcal{T}}}$ where $p_{\mathcal{S}}$ and $p_{\mathcal{T}}$ are the respective marginal distributions. The function $c:X\times Y\rightarrow \mathbb{R}^+$ serves as a cost function, such as the Euclidean distance measured by the $\ell_2$-norm.
In general, the Wasserstein Distance  does not have a closed-form solution. However, when dealing with one-dimensional distributions, a closed-form solution exists and is expressed as follows:
\begin{equation}
W(p_{\mathcal{S}},p_{\mathcal{T}})= \int_{0}^1 c(P_{\mathcal{S}}^{-1}(\tau),P_{\mathcal{T}}^{-1}(\tau))d\tau.
\label{eq:oneD}
\end{equation} 
Here, the cumulative distributions of the one-dimensional distributions $p_{\mathcal{S}}$ and $p_{\mathcal{T}}$ are denoted as $P_{\mathcal{S}}$ and $P_{\mathcal{T}}$, respectively. This closed-form solution has inspired the introduction of SWD, which utilizes the slice sampling technique proposed by Neal \cite{neal2003slice}. The concept of slicing involves selecting a one-dimensional subspace and projecting the two $d$-dimensional distributions onto this subspace. This projection generates the marginal one-dimensional distributions, based on which SWD is defined. SWD is computed by integrating the one-dimensional WD distances between the marginal distributions over all possible one-dimensional projection subspaces.
For a distribution $p_{\mathcal{S}}$, a one-dimensional slice is defined as:
\begin{equation}
\mathcal{R}p_\mathcal{S}(t;\bm{\gamma})=\int_{\mathcal{S}^{d-1}} p_\mathcal{S}(\bm{x})\bm{\delta}(t-\langle\bm{\gamma}, \bm{x}\rangle)d\bm{x},
\label{eq:radon}
\end{equation}
where, the Kronecker delta function $\bm{\delta}(\cdot)$ is used for projection using the inner dot product  $\langle \cdot ,\cdot\rangle$. The unit sphere in a $d$-dimensional space is represented by $\mathbb{S}^{d-1}$, and $\bm{\gamma}$ represents the one-dimensional projection direction. Note that the projection direction is a variable.
The SWD   is defined as an integral over all possible projections, which can be expressed as follows:
\begin{eqnarray}
SW(p_\mathcal{S},p_\mathcal{T})=   \int_{\mathbb{S}^{d-1}} W(\mathcal{R} p_\mathcal{S}(\cdot;\gamma),\mathcal{R} p_\mathcal{T}(\cdot;\gamma))d\gamma.
\label{eq:radonSWDdistance}
\end{eqnarray}
 
  SWD offers a significant advantage over WD    because unlike WD, the integrand of Eq. \eqref{eq:radonSWDdistance} has a closed-form solution when $\gamma$ is known. What remains is how to compute the integral.
To compute the integral in Equation \eqref{eq:radonSWDdistance}, we can employ a Monte Carlo style integration approach. For this purpose, we sample the projection subspace $\bm{\gamma}$ from a uniform distribution defined over the unit sphere. Then, we calculate the 1-dimensional WD and approximate the integral in Eq. \eqref{eq:radonSWDdistance} by taking the arithmetic mean over a sufficiently large number of random projections. This Monte Carlo approach allows us to numerically estimate the SWD:
\begin{equation}
\hat{D}(p_\mathcal{S},p_\mathcal{T})\approx \frac{1}{L}\sum_{l=1}^L \sum_{i=1}^M| \langle\gamma_l, \phi(\bm{x}_{s_l[i]}^\mathcal{S}\rangle)- \langle\gamma_l, \phi(\bm{x}_{t_l[i]}^\mathcal{T})\rangle|^2
\label{eq:SWDempirical}
\end{equation}
where $\gamma_l\in\mathbb{S}^{f-1}$ is a random sample drawn uniformly from the unit $f$-dimensional ball $\mathbb{S}^{d-1}$. For each domain, we sort the indices of ${\gamma_l\cdot\phi(\bm{x}i)}{i=1}^M$, denoted as $s_l[i]$ and $t_l[i]$ for the source and target domains respectively. In We adopt the empirical computation of the SWD using Eq.~\eqref{eq:SWDempirical} which has been used to address UDA~\cite{lee2019sliced,rostami2021domain}. 
 The choice of SWD as a metric in our work is motivated by several reasons. Firstly, as previously discussed, SWD is a suitable metric for optimizing deep learning models. Secondly, it has a closed-form solution that enables efficient computation. Lastly, its empirical version can also be computed efficiently using Eq.~\eqref{eq:SWDempirical}.

Numerous UDA methods have been developed based on variations of Eq.\eqref{eq:smallmainPrMatch}, utilizing different probability metrics and additional regularization techniques to enforce specific properties. In our work, we aim to improve upon previous approaches that solve Eq.\eqref{eq:smallmainPrMatch} by inducing larger margins between the learned class clusters in the embedding space. This margin enlargement helps mitigate the negative impact of domain shift in the target domain by improving model generalizability. Figure~\ref{figMUDA:1} provides a conceptual visualization of the positive effect achieved by creating larger interclass margins, which leads to more compact feature representations.

\end{document}